\documentclass[runningheads]{llncs}
\usepackage{graphicx}
\usepackage{url}
\usepackage{comment}
\usepackage{mathtools}
\usepackage{color}
\usepackage{booktabs}
\usepackage{multirow}
\usepackage{hyperref}
\usepackage{url}
\usepackage{comment}
\usepackage{amsmath,amssymb} 
\usepackage{subcaption}
\usepackage{mathtools}
\usepackage{color}
\usepackage{hyperref}
\hypersetup{
    colorlinks=true,
    linkcolor=blue,
    filecolor=magenta,      
    urlcolor=cyan,
}

\newcommand{\rightarrowgtr}{\ensuremath{\mathrlap{\;\,\rightarrow}>}}

\begin{document}

\pagestyle{headings}
\mainmatter
\title{Image segmentation via Cellular Automata}
\author{Mark Sandler\inst{1} \and
Andrey Zhmoginov\inst{1}\and
Liangcheng Luo\inst{1}\and
Alexander Mordvintsev\inst{1} \and
Ettore Randazzo \inst{1} \and
Blaise Ag\'{u}era y Arcas \inst{1}
}
\institute{Google AI \\ \email{\{sandler,azhmogin,luolc,moralex,etr,blaisea\}@google.com}\\}
\maketitle

\begin{abstract}
      In this paper, we propose a new approach for building cellular automata to solve real-world segmentation problems.     We design and train a cellular automaton that can successfully segment high-resolution images.    We consider a colony that densely inhabits the pixel grid, and all cells are governed by a randomized update that uses the current state, the color, and the state of the $3\times 3$ neighborhood. The space of possible rules is defined by a small neural network.    The update rule is applied repeatedly in parallel to a large random subset of cells and after convergence is used to produce segmentation masks that are then back-propagated to learn the optimal update rules using standard gradient descent methods. 
    
    We demonstrate that such models can be learned efficiently with only limited trajectory length    and that they show remarkable ability to organize the information to produce a globally consistent segmentation result, using only local information exchange. 
    
    From a practical perspective, our approach allows us to build  efficient models -- our smallest automaton uses less than 10,000 parameters to solve complex segmentation tasks.
   
\end{abstract}

\section{Introduction}
    Cellular automata popularized by Conway's Game Of Life~\cite{gameoflife} and Stephen Wolfram's book ``A New Kind of Science'' \cite{newkindofscience} have been extensively studied and applied in various fields from physics and cryptography to dynamical systems and biology of multicellular organisms.

    The core component of cellular automata (CA) is a local update rule defining a system transition over a single discrete time step.
    One of the most amazing properties, is that  even the most straightforward time-independent update rules can result in very diverse and complex system behaviors. For example it has been known for a long time Conway's Game of Life is Turing complete~\cite{lifeisturingcomplete}.  In this work we turn away from hand-designed automata and instead learn the update rule using SGD. An alternative view of cellular automata is to think of cellular automata as a recurrent neural network. Such approach have been thoroughly explored as a model for sequence processing with applications ranging from natural language processing to reinforcement learning.   Most widely used recurrent cell types including RNNs, GRUs, LSTMs, and based on the iterative data processing with transformations that share their parameters.
    In recent years, recurrent architectures have also been explored in application to computer vision tasks ranging from object segmentation and salient region discovery to image super-resolution (see Section~\ref{sec:related}).
    The critical difference is that the vast majority of the prior work,  use a small number of recurrent iterations and allow for long-range interactions via reduced resolution of deeper layers, thus departing from a traditional setting of a cellular automaton with local update rules.  In this paper, we propose a novel architecture that uses a single update rule to iteratively arrive at the correct solution. 

    Our contributions are twofold.
    First, we demonstrate that a recurrent model with only {\em local} or quasi-local interactions (where we allow the usage of global statistics), can be trained to perform well in a highly nontrivial context-aware task of image segmentation.
    Secondly, we propose a set of techniques that allow us to limit the number of unroll steps and make it possible to stabilize the training of cellular automata over long periods.   Perhaps what is even more interesting, with our techniques, trained models can adapt to a change of the underlying image, at least partially reusing information obtained while processing the original image. These models can thus be directly applicable to video processing.

    The paper is organized as follows. In the next section we overview the related work. In section \ref{sec:design} we provide detailed design of our automata, including some of the technical decisions we had to make to stabilize the training. Finally in section \ref{sec:experiments} we present our experimental results. 

\section{Related work}
\label{sec:related}

\paragraph{Recurrent models in computer vision applications.}
    Recurrent models so widely used in NLP have also been applied to computer vision tasks.
    Some work like \cite{DBLP:conf/cvpr/VisinRCMCKBC16} revolved around direct applications of sequence models to image processing.
    Another common theme is the idea that the same transformation can effectively be applied to different image scales, thus resulting in a recurrent CNN architecture.
    This approach has previously been applied to object classification \cite{DBLP:conf/cvpr/LiangH15,DBLP:journals/corr/AlomHYT17}, image super-resolution \cite{DBLP:conf/cvpr/KimLL16}, label attribution \cite{DBLP:conf/icml/PinheiroC14,DBLP:conf/nips/LiangHZ15} and salient region detection \cite{DBLP:journals/pami/WangWLZR19}.
    Similarly, in \cite{DBLP:journals/corr/abs-1802-06955} authors explore recurrent convolutional layers as individual elements in a larger U-Net style network.
    In \cite{DBLP:conf/iccv/0001JRVSDHT15}, recurrent CNN architecture emerges when approaching the label attribution problem using conditional random fields.

\paragraph{Hybrid models.}
    Another natural application of recurrent convolutional models is in video processing or, more generally, in situations when the visual information has some form of temporal consistency.
    For example, in \cite{DBLP:journals/tmi/QinSCPHR19}, recurrent CNNs are used for magnetic resonance imaging processing.
    In some approaches (see for example, \cite{DBLP:conf/cvpr/KuenWW16,DBLP:conf/wacv/ValipourSJR17,DBLP:conf/cvpr/McLaughlinRM16,DBLP:journals/spl/ZuoFBL17,DBLP:conf/cvpr/TrigeorgisSNAZ16,DBLP:journals/corr/abs-1903-10172,DBLP:conf/cvpr/LiuZ18}), single image representations produced by a CNN are fed into an RNN or LSTM-based recurrent model.
    A hybrid approach, where recurrent architectures are used for both the base CNN model and the processing of temporal information, can, for example, be found in \cite{DBLP:conf/cvpr/LeeO16}.

\paragraph{Back to cellular automata.}
    While all aforementioned applications inherit the core idea of recurrence from cellular automata, most depart from this simple model by introducing (explicitly or implicitly) non-local information exchange.
    Several recent papers including \cite{DBLP:journals/corr/abs-1809-02942} turned back to the setting of the cellular automaton with the learned {\em local} recurrent update rules.
    Such systems have recently been explored in a range of topics from studying self-organization \cite{selforg} to proposing a general computational platform \cite{DBLP:journals/corr/KaiserS15,DBLP:journals/corr/FreivaldsL17}. 
    
    The usage of cellular automata for image  segmentation was also explored in earlier papers  \cite{ParameterizedCellularAutomataInSegmentation,AvenuesForCellularAutomataInSegmentation}. However the automaton considered in those works had a simple hand-designed parameterization used to detect similar regions,
    in a way similar to flood-fill type of  algorithms. Here, in contrast we {\em learn} an  automaton that can arrive to a very complex update rules. 

\section{Detailed Cellular Automaton Design}
\label{sec:design}

\def\R{\mathbf{R}}
\def\SF{{\cal S}}
\def\state{C}
\def\image{{\cal I}}
\def\neighborhood{{\cal N}}
\def\height{h}
\def\width{w}
\def\softmax{{\rm softmax}}
\def\css{d}

\begin{figure}[t]
    \centering
    \begin{subfigure}[b]{0.97\textwidth}
        \centering
        \includegraphics[width=\textwidth]{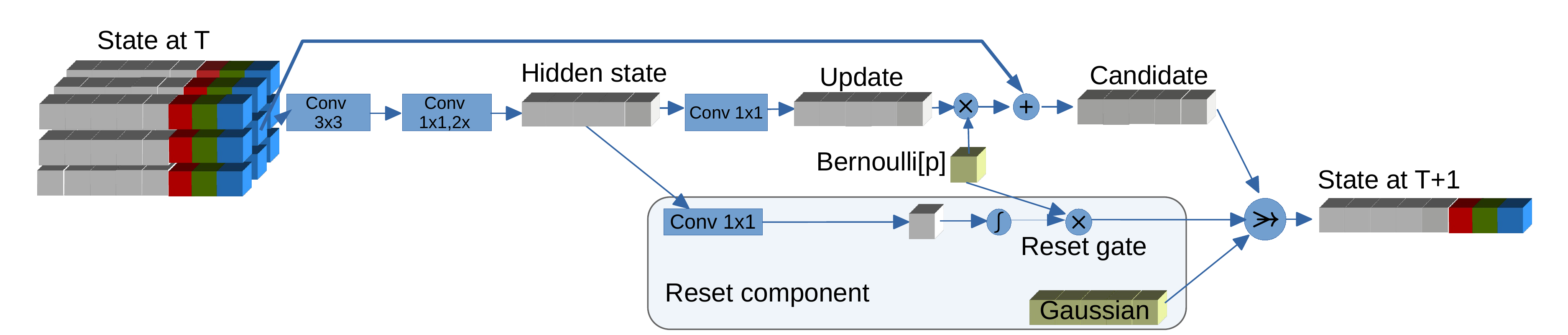} 
    \end{subfigure} 
    \caption{A diagram of a single cell update rule.
    For non-resettable cells, the \emph{reset component} is removed and the candidate always becomes the new cell state.
    The Bernoulli variable randomly disables changes to the cell to simulate asynchronous evolution of different cells. 
    For resettable cells, the next state is a mixture (denoted by $\rightarrowgtr\;$) of random noise and the candidate sate, with the reset gate determining the relative weight of each component.
    }
    \label{fig:cell_update_rule}
\end{figure}

    Following a common cellular automaton design similar for example to the one used in \cite{selforg}, we consider a grid space of size $h \times w$ inhabited by cells which can change their state based on the state of their closest neighbors and the state of the {\em environment} around them.
    In the following, we assume that each cell has a mutable state $s \in \R^{\css}$.
    Then, for image processing tasks, we define the environment of a cell by associating it with color channels of the corresponding image pixel.
    More formally, a single step of the evolution of a cell state $s$ corresponding to a pixel $i$ is assumed to be given by:
    $$
    s \leftarrow \SF(\neighborhood(s), \neighborhood(i)),
    $$ 
    where all cell states are assumed to be updated simultaneously, $\neighborhood$ denotes an immediate neighborhood of a cell or a pixel and $\SF: \R^{n{\css}} \times \R^{nk} \rightarrow \R^{\css}$.
    Here $k$ is the size of the environment state ($k=3$ for a standard RGB image) and $n$ is the size of the neighborhood each cell is allowed to read. 
    In this paper, we use the neighborhood of size $9$ -- that is $8$ immediate neighbors plus the cell itself.
    The update rule $\SF$ is a state transition that we {\em learn} with the goal of training the resulting cellular automaton to execute a specific task.
    We start by using update rules of the form $\SF(s)=U(s) + s$, where $U(s)$ is a funciton that takes the neighborhood and produce state update. The space of functions we consider is 3-layer neural network consisting of single $3\times 3$ convolution and $1\times 1$. This update rule was inspired by residual blocks \cite{ResNet,ResNext2016}, however
    the critical difference is that rule is \emph{the same}  throughout the entire model.  In experimental section we also explore variants of cell-updates without residual connections. In the latter case the update rule is simply $U(s)$.

    We use $\state_t$ to denote the entire $\height \times \width \times d$ state at a given time $t$.
    The progression over $t$ time steps is then equivalent to a repeated application of the transformation $\SF$.
    That is $$ \state_t = \SF^t(\state_0,\image) = \underbrace{\left[\SF \circ \ldots \circ \SF\right]}_{t \, \text{times}}(\state_0,\image), $$
    where the same image $\image$ is provided at every step.
    For our experiments, we will use randomly initialized initial state $\state_0$ sampled from standard Gaussian distribution.

    We use a simple multi-layer neural network to compute the state update on each step.
    That is
    $$
      \SF(s, i) = [L_k \circ \dots \circ L_1](\neighborhood(s), \neighborhood(i))
    $$
    where each $L_i$ is simply a fully connected layer with non-linearity that maps the state to a hidden representation.
    Since we use a dense colony, from a practical perspective, this can be efficiently implemented by applying the standard convolutional operator to the entire $\height \times \width \times (d+k)$ state.  
    We note that in our architecture, only the first layer $L_1$ uses the information about the pixel neighbors.
    This highlights the fact that only minimal interaction between neighbors is required.
    The remaining operations are only using the current state.
    In our implementation, this translates to using $1\times 1$ convolutions for all the layers except the first one. 
    
    In \cite{selforg} it was suggested to further limit spatial interaction to a directional gradient.
    Instead, here we use learned spatial relationships for simplicity.
    Interestingly, the {\em actual} spatial relationship appears to be of only modest importance, and even using only random spatial depthwise convolution, the cellular automata can still perform the task fairly well.
    See section \ref{sec:ablation} for more details.
    
    In its simplest variation, the output of the neural network is considered to be a proposed state update.
    The cell state is then incremented by this proposed update with a fixed probability $p$ sampled independently for each pixel.
    This is done in a way similar to \cite{selforg} to ensure the stability of the colony.
    In our experiments, we always use per-step update probability $p=0.5$.
    In Section \ref{sec:resettable} we generalize this function to include the ability to control self-reset of the state.

\subsection{Objective and training}
\def\loss{L}
    Now that we defined the cellular automaton, how do we find the update rule $\SF$?
    First of all, it is unrealistic to expect the colony to arrive at a good solution in one step.
    Instead, we pick a constant $T$, the {\em target} number of steps, after which we expect the colony to solve the problem.
    We can then use a standard cross-entropy loss function that is computed at every step of the colony once the colony has observed the image for longer than the initial threshold of $T_0=T/3$. 
    For each spatial location $i$ with state $s$ we can compute raw predictions $\hat{y} = \softmax(W s_t)$ using the learned weights $W$ and use the cross-entropy loss:
    $$
    \loss(\state_t, y) = -\mathbb{E}_{i} \sum_{j}  y_j \log [\softmax(W \state_{i,t})]_j,
    $$
    where the outer expectation is over all spatial locations and the inner summation is over all classes. 
    The most natural training procedure widely used for training recurrent architectures is to unroll the cell state for $T$ steps and use standard gradient descent with cell state $s$ up to the number of steps and propagate the gradient through it.
    However, this approach results in a cellular colony overfitting to that particular number of steps $T$, and the prediction deteriorates before and after this step count.
    Even more crucially, the number of steps needed to produce good predictions in our experiments is fairly large ($>20$).
    Training such unrolled cellular automaton requires large amounts of memory.
    In the remainder of this section, we describe additional techniques that allow us to significantly reduce the length of unrolls and improve overall model performance.

\subsection{Mini-unrolls}
    \label{sec:preserve-state} \label{sec:mini-unrolls}
    Naive training of a cellular automaton with standard SGD looks very much like training a regular deep neural network.
    In particular, it requires (a) back-propagating through all $T$ unrolled steps followed by (b) switching the images and resetting the cell state to process the following batch. 
    As seen in Figure~\ref{fig:evolution-of-iou-over-long-run}, this approach to training a model leads to unstable outcomes where cellular automata diverge if allowed to proceed over a large number of steps. 
    We therefore modify a single step of SGD by running it for only $K \ll T$ steps and then {\em reusing} a fraction of images and states after each mini-unroll.
    The probability $p$ of reusing the state is chosen in such a way that only the fractions $p_i$ and $p_s$ of images and states correspondingly get reset before reaching the full-unroll target $T$.
    This achieves the following objectives: 
    \begin{enumerate}
        \item Reset the image, but keep the state -- allows the colony to learn to update image state preventing it from being frozen;
        \item Keep the image and keep the state -- allows the colony to stabilize the prediction for a larger number of steps;
        \item Reset the state, either keep or reset the image -- standard SGD-like step where we ensure that a colony can converge to a correct solution starting from random initialization.
    \end{enumerate}
    To achieve the target reset probability $p$ at step $T$, we use a per-unroll reset probability of $p^{K/T}$.
    This change allows us to scale the complexity and size of the colony without running out of memory, but also makes the colony capable of adapting to image changes.
    However, as shown in Figure~\ref{fig:evolution-of-iou-over-long-run}, the accuracy still deteriorates significantly in the long run.
    In the next section, we describe the approach that allows the colony to adapt without degradation continuously.

\subsection{Reset gates for cell states}
\label{sec:resettable}
    Cellular automaton design described in Section~\ref{sec:design} converges when looked at the narrow task of predicting image after a fixed number of steps.
    In practice, we observed that the cell states tended to converge to quasi-stable equilibrium that slowly deteriorated over time.
    For instance, in Figure~\ref{fig:evolution-of-iou-over-long-run} the colony trained using standard SGD diverged after several hundred steps.
    Models trained this way also did not properly respond to a change of the underlying image.
    This latter property is important, for example, for video processing, but we observed that once the input image was updated, the cell state of the trained model slowly deteriorates. One such example is shown in Figure \ref{fig:evolution-of-iou-with-shifts} where slight shifts to image lead to significant performance degradation.
    More formally, if $\state_0$ is some initial cell state and $\image_0$ is the input image, then empirically we observe that once the colony converges to a stable state $S_{\image_0} = \SF^{*}(\state_0, \image_0)$
    $$
    \SF^{\tau}(s_{\image_0}, J) \sim s_{\image_0}
    $$
    for an arbitrary image $J$ and large values of $\tau\in \mathbb{N}$.
    Note that we use $\sim$ here to indicate that predictions for the states on both sides approximately match, while the states themselves may be different (and change over time).

    This behavior is undesirable if we want the colony to properly evolve in response to a changing input.
    To address this, we introduce a ``reset gate'' reminiscent to that of GRU~\cite{GRULearningToForget} recurrent units, where a cell state can be reset based on its state.
    However one difference is that we use continuous source of randomness rather than resetting the state to zero.
    The new resettable state update $\SF_r$ function now looks as follows.
    $$
    \begin{array}{rcl}
    r(s) &= &\sigma[W_R H(s)] \\ 
    \SF_r(s) &= &r(s) z + (U(s) + s) (1 - r(s))
    \end{array}
    $$
    where $z\in R^{\css}$ is a random normal variable that provides independent random noise for each cell, $H(S)$ is a hidden state -the last layer before update, and $W_R$ is a trainable $\css \times 1$ matrix describing the transformation from the hidden state to a scalar indicating whether the cell should reset. 
    The diagram of the full resettable cell is shown in Figure~\ref{fig:cell_update_rule}.
    

\subsection{Cellular automata for high-resolution images}
    \label{sec:high_resolution_idea}
    Applying our method as described to high-resolution images would require excessive computational resources and lead to a very large memory footprint during training.
    While it is a departure from our previous design, we found that the most accurate and scalable approach to dealing with high-resolution images is to use CA state resolution that is smaller than the image resolution.
    In particular, we used convolutions and transpose convolutions with a finite stride to: (a) reduce the input image size to make it compatible with the CA state resolution and (b) transform the CA state into a high-resolution segmentation map.
    Experimental results with this approach are presented in Section~\ref{sec:high-res}.

\subsection{State Normalization}
\label{sec:state_normalization}
    Our CA models converge well even when no cell state normalization is used.
    However, we generally observed that normalization further stabilized optimization.
    We experimented with four types of normalization: batch-normalization \cite{batchnorm} in the training mode and instance normalization \cite{instancenorm}, channel normalization~\cite{ba2016layer} and no normalization at all. 
    In our experiments, we found that the normalization methods often perform comparably, however have different limitations in practice. For instance for batch normalization we had to rely on the live mini-batch statistics even in evaluation mode to avoid state-shift caused by the cell state evolution. 

\section{Neural networks vs. Cellular Automata}
    As described in \ref{sec:design}, the proposed cellular automata utilize the same type of computation as regular neural networks. 
    However, the critical difference is how the information is propagated.
    In standard neural network architectures, a common intuitive picture implies that the deeper layers specialize in higher-level concepts using a different set of learned variables.
    In the case of a cellular automaton, the deeper layers are actually the same as earlier layers -- and they all just evolve the state on the grid. 
    
    An alternative view is to treat these cellular models as recurrent convolutional models~\cite{DBLP:conf/iclr/SavareseM19,DBLP:journals/corr/abs-1802-06955,DBLP:conf/icml/PinheiroC14}, where the output of each sequence of loops is fed into the model.
    However, such intuition does not capture the local-organization aspect.
    Indeed, the state cell on each step is only ever affected by the immediate neighbors.
    Thus, it is imperative for the model to propagate local information across many steps to achieve spatial consistency.
    
    A variant of such local propagation is also exploited in {\em Isometric Networks} \cite{isometric}, where all layers have fixed spatial dimensions, and no layers with stride are used.
    This, similarly to present work forced the information propagation locality.
    However, in~\cite{isometric}, there was no recurrence that allowed iterative incrementality of computation that cellular automata allow. 

\section{Experiments}
\label{sec:experiments}
    In this section we present qualitative and quantitative results of applying our cellular automata to the Oxford Cats and Dogs Dataset \cite{oxford-pets}. 
    This dataset contains about 3400 training and 3400 evaluation samples.
    The goal of our cellular automata was to segment the image pixels into ``background'', ``object'' and ``object boundary'' regions.
    
    First, we describe our architectural setup.
    Unless noted otherwise, the length of mini-unroll of $10$ steps and the target prediction goal of $40$ steps in all our experiments.
    Each cell is described by a $48$-dimensional state and contains 4 hidden layers.
    Inside the sequence of hidden layers, we use a state of size $64$, i.e., the first convolutional operator uses transforms from 48 to $64$ channels.     The first layer in these sequence is a $3\times 3$ convolution and the remaining layers are all $1\times 1$ convolutions.
    The prediction is obtained by applying an additional $1\times 1$ convolutional layer to the automaton state to get \emph{num-classes} prediction vector for each pixel.
    We use RELU for all hidden layers, and use linear activations as is standard for logit predictions. We use instance normalization~\cite{instancenorm} for most of our experiments and standard Adam optimizer~\cite{Kingma2014AdamAM} with the learning rate $3\cdot 10^{-4}$ and a batch size of $32$.  We train all our runs for 1 million steps.    
    \begin{figure}[t]
        \centering
        \begin{subfigure}[b]{0.99\textwidth}
            \centering
            \includegraphics[width=\textwidth]{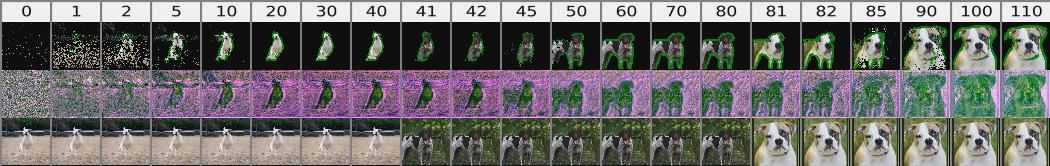} 
        \end{subfigure} 
        \caption{\small{Evolution of the colony with resettable cells where the underlying images change every 40 steps. The first row shows the prediction.  The second row is a projection of the hidden state on RGB. Note how the predicted shape adopts to the image.}}
    \end{figure}
    
\subsection{Qualitative Experiments}
    \begin{figure}[t]
        \centering
         \begin{subfigure}[b]{0.97\textwidth}
            \centering
            \includegraphics[width=\textwidth]{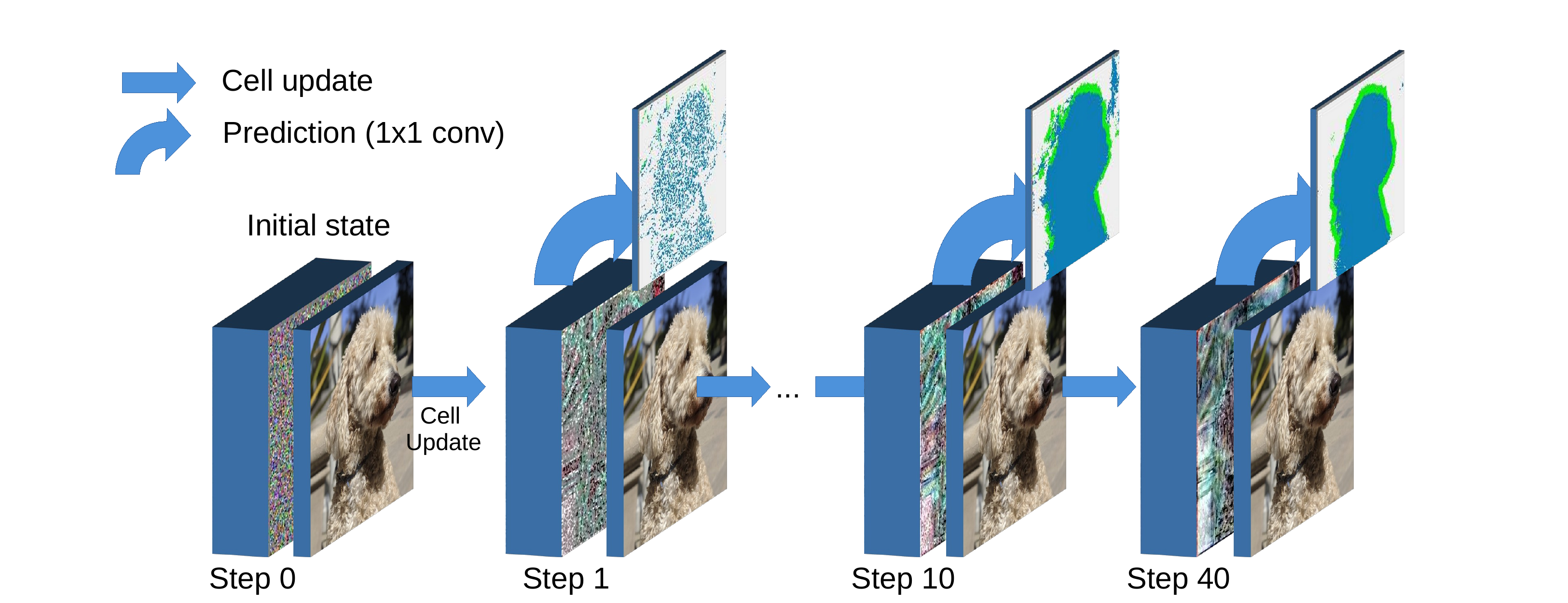}
        \end{subfigure} 
        \caption{Evolution of a cell colony for segmentation. Each step apples the same update rule (see Figure~\ref{fig:cell_update_rule}) to each cell independently.}
        \label{fig:colony_progression}
    \end{figure}
\begin{figure}
\centering
\begin{minipage}{0.99\textwidth}
\begin{subfigure}[b]{0.19\textwidth}
         \centering
    \includegraphics[width=\textwidth]{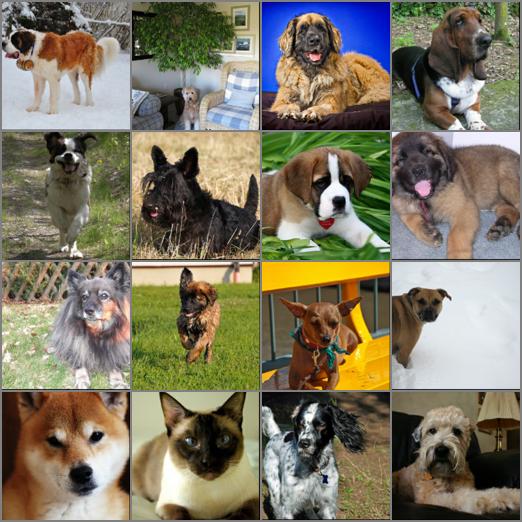}
    \caption{Original}
    \label{fig:original-sample}
\end{subfigure}
\begin{subfigure}[b]{0.19\textwidth}
         \centering
    \includegraphics[width=\textwidth]{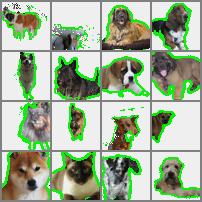}
    \caption{\small{d=48}}
\end{subfigure}
 \begin{subfigure}[b]{0.19\textwidth}
         \centering         
    \includegraphics[width=\textwidth]{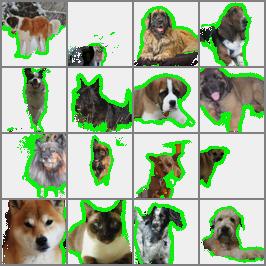}
    \caption{d=64}
\end{subfigure}
\begin{subfigure}[b]{0.19\textwidth}
    \centering
    \includegraphics[width=\textwidth]{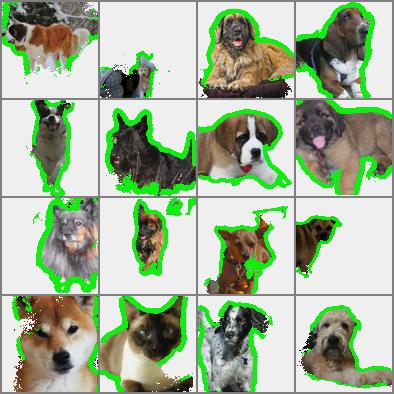}
    \caption{d=96}
\end{subfigure}
\begin{subfigure}[b]{0.19\textwidth}
    \centering
    \includegraphics[width=\textwidth]{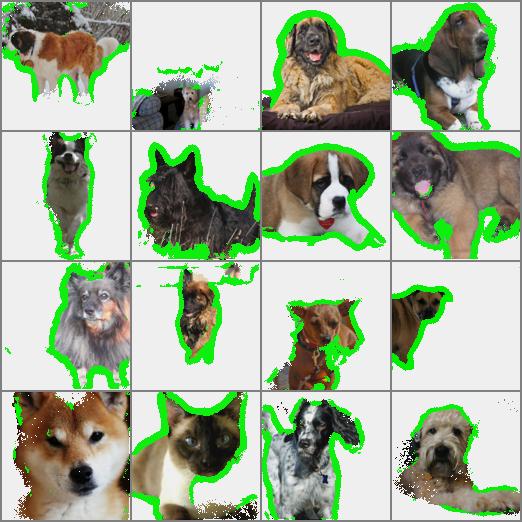}    
    \caption{d=128}
\end{subfigure}
\caption{Samples of predictions of colonies trained to segment at different resolutions $d$. All predictions are after 40 steps.}
\label{fig:different-resolution}
\end{minipage}
%
\begin{minipage}{0.99\textwidth}
\vspace{0.5cm}
\begin{subfigure}[b]{0.19\textwidth}
         \centering
    \includegraphics[width=\textwidth]{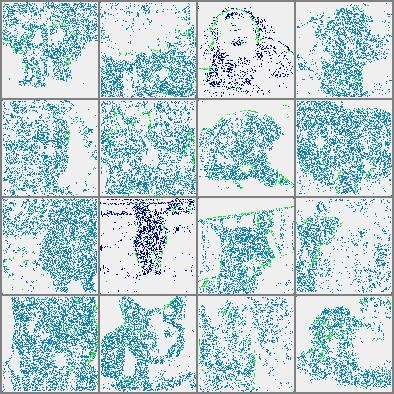}
    \caption{1 step}
\end{subfigure} 
\begin{subfigure}[b]{0.19\textwidth}
         \centering
    \includegraphics[width=\textwidth]{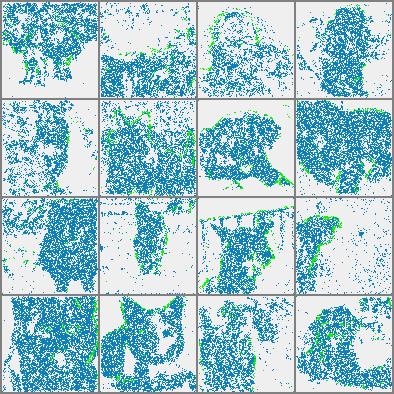}
    \caption{2 steps}
\end{subfigure} 
 \begin{subfigure}[b]{0.19\textwidth}
         \centering
    \includegraphics[width=\textwidth]{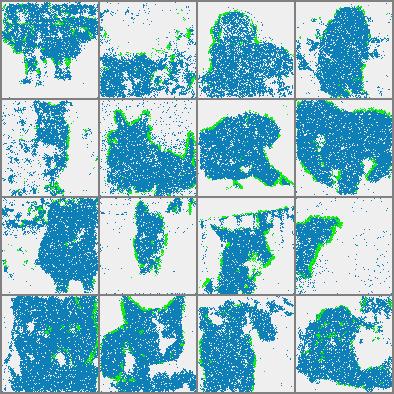}
    \caption{4 steps}
\end{subfigure} 
\begin{subfigure}[b]{0.19\textwidth}
    \centering
    \includegraphics[width=\textwidth]{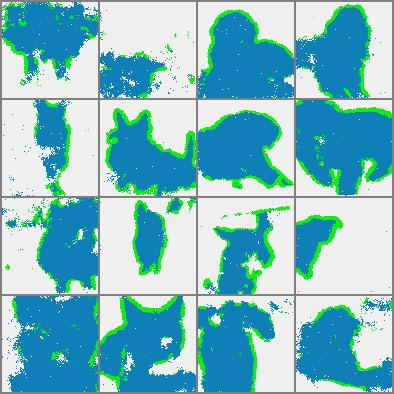}    
    \caption{10 steps}
\end{subfigure}     
\begin{subfigure}[b]{0.19\textwidth}
    \centering
    \includegraphics[width=\textwidth]{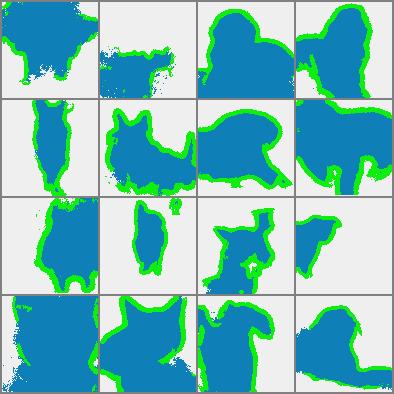}    
    \caption{40 steps}
\end{subfigure}     
\caption{Intermediate predictions after different number of steps. }
\end{minipage}
\end{figure}

    We show prediction results of a trained colony on a validation sample of 16 images in Figure~\ref{fig:colony_progression}.
    The original images are shown in Figure~\ref{fig:original-sample}.
    As we can see in Figure~\ref{fig:colony_progression}, the prediction effectively starts close to random at step $1$.
    By step $10$ model predictions take the correct shape and the model uses the remaining steps to refine the prediction.

\subsection{Ability to adapt to image changes (shifts and resets)}
\begin{figure}
\centering
 \begin{subfigure}[b]{0.31\textwidth}
         \centering
    \includegraphics[width=\textwidth]{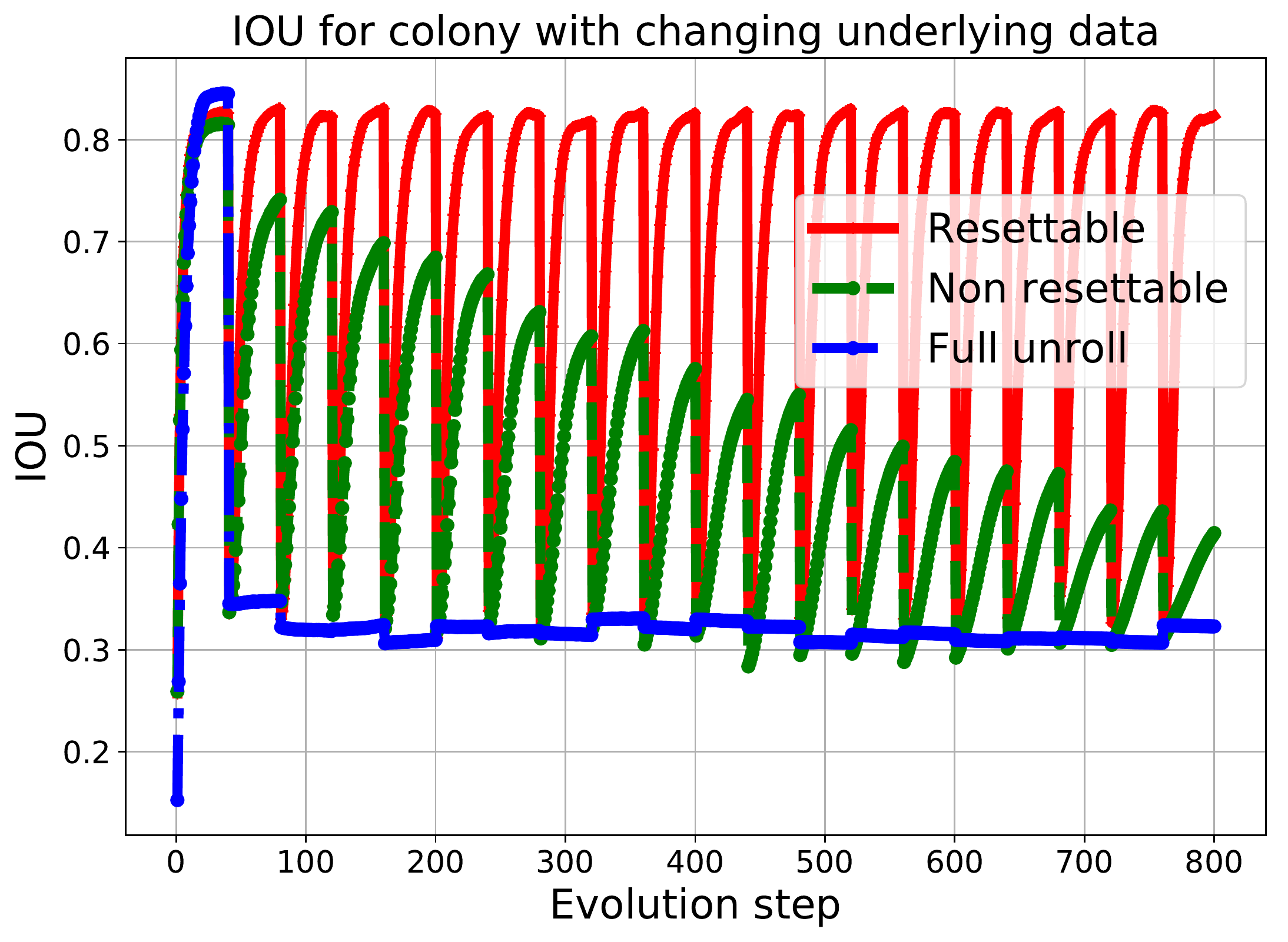}
    \caption{Images change every 40 steps.}
    \label{fig:images-change-every-40}
\end{subfigure} 
\begin{subfigure}[b]{0.31\textwidth}
         \centering
    \includegraphics[width=\textwidth]{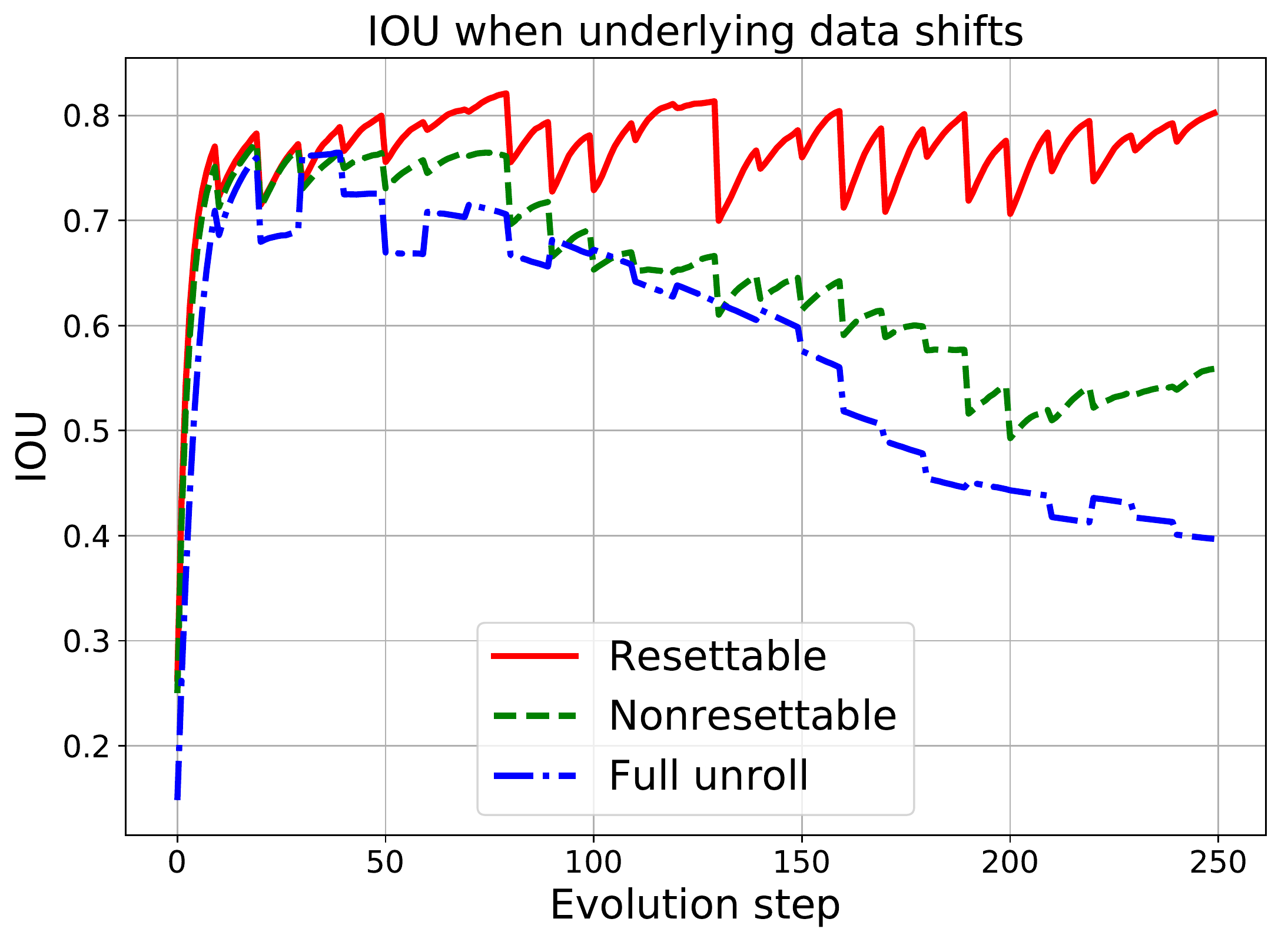}
    \caption{Images are randomly shifted every 10 steps. }
    \label{fig:evolution-of-iou-with-shifts}
\end{subfigure} 
\begin{subfigure}[b]{0.31\textwidth}
         \centering
    \includegraphics[width=\textwidth]{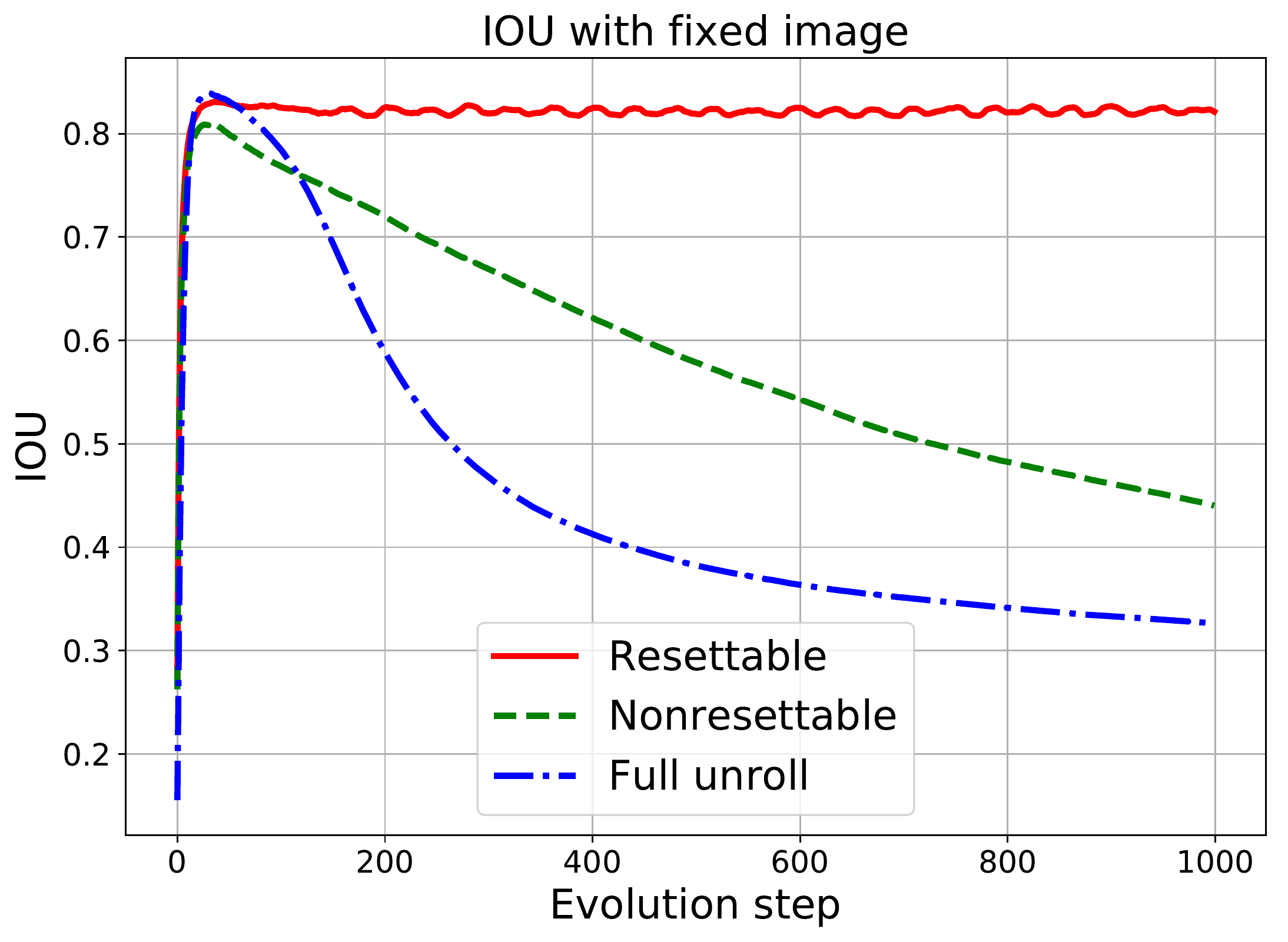}    
    \caption{Evolution of colony over stable image.}
    \label{fig:evolution-of-iou-over-long-run}
\end{subfigure} 
\caption{IOU performance when underlying image changes for different types of models. The resettable and non-resettable models
preserve the state and the images with probability 0.5 and use short unroll of size 10. Full unroll model uses classical SGD with full
state and image resets. Note how introducing reset gate essentially stabilized the performance over a large number of evolution steps.}
\label{fig:accuracy-decay-per-step}
\end{figure}

    In Figure~\ref{fig:accuracy-decay-per-step} we show how different models adapt to different types of image changes.
    The full-unroll model is essentially a model that is trained like a regular neural network with random sample at each batch and a full reset of the state.
    As one can see, the model rapidly reaches its highest accuracy, but then rapidly degrades to a random prediction.
    On the other hand, the non-resettable model that maintains some components of the state with images that change as described in Section~\ref{sec:preserve-state} shows a much more gradual degradation.
    However, after a few image changes it too degrades severely.
    Finally, after soft reset gates are introduced, the model becomes fully stable.

\begin{figure}[t]
\centering
 \begin{minipage}[t]{0.32\textwidth}
         \centering         
    \includegraphics[width=\textwidth]{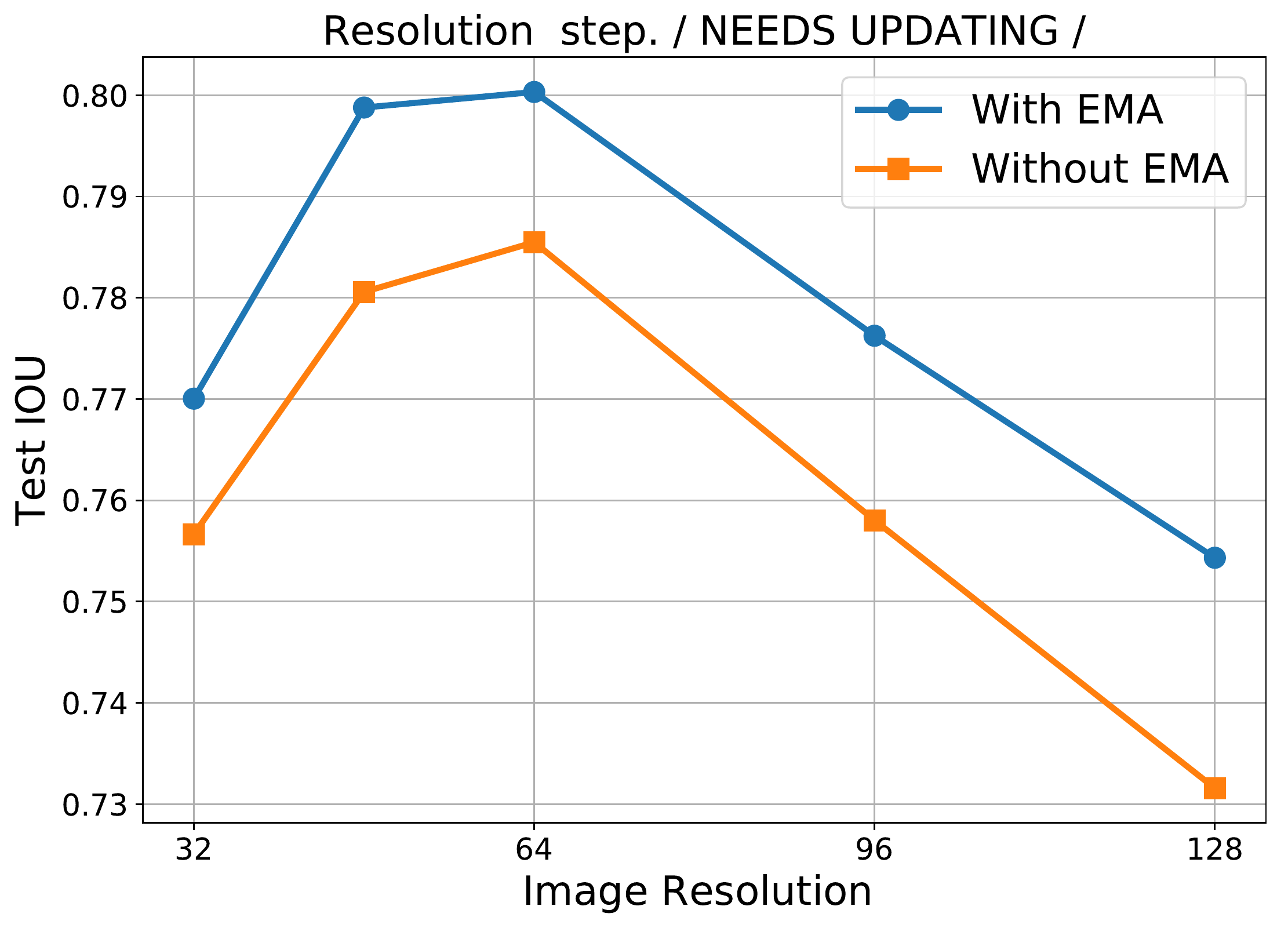}    
    \caption{\small{Image size vs. IOU}}
    \label{fig:image_size_vs_iou}
\end{minipage} 
\begin{minipage}[t]{0.32\textwidth}
         \centering
    \includegraphics[width=\textwidth]{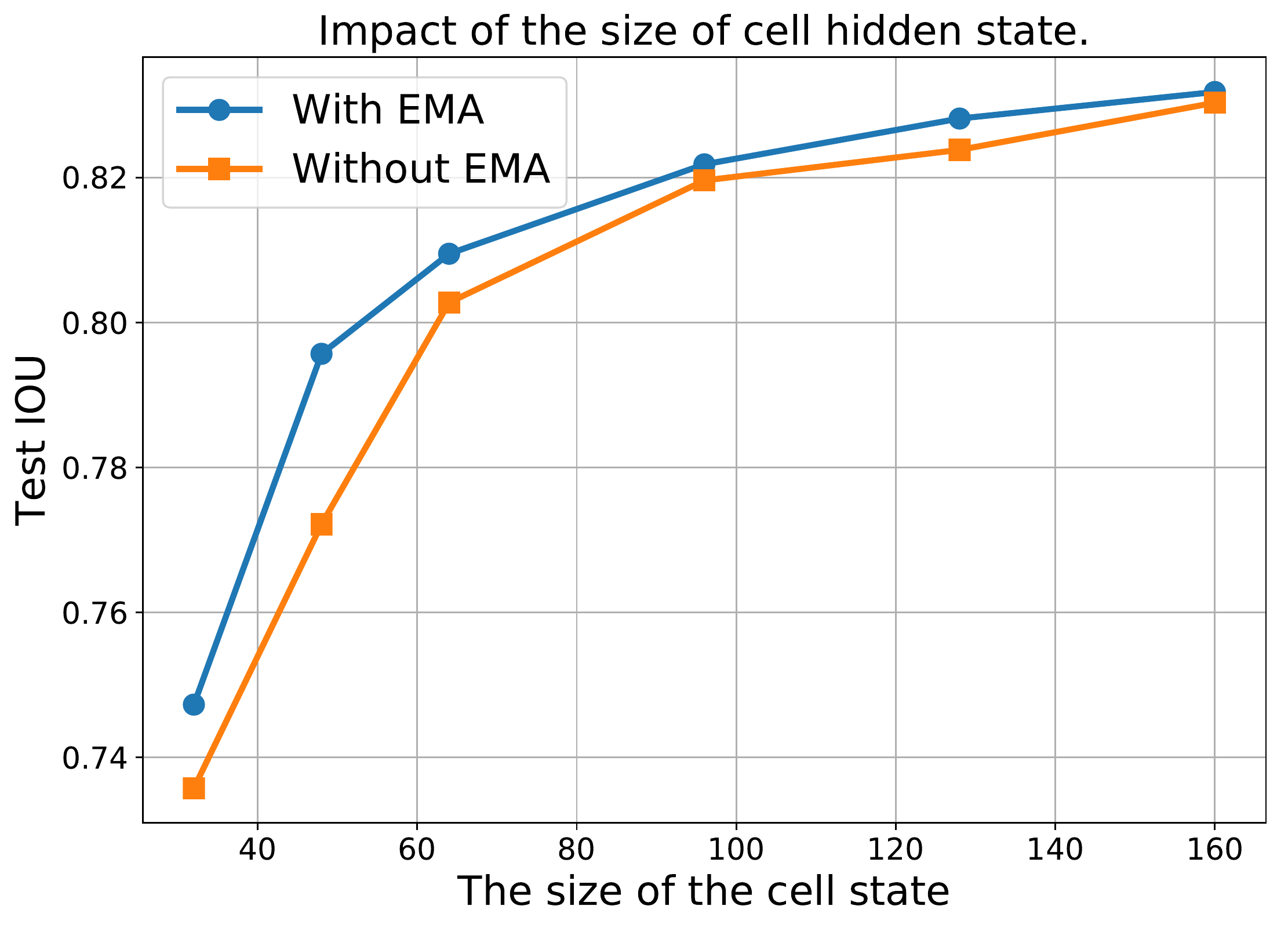}        
    \caption{\small{Cell state vs. IOU}}
    \label{figure:cell_state_vs_iou}
\end{minipage}     
\begin{minipage}[t]{0.32\textwidth}
     \centering
         \centering
        \includegraphics[width=\textwidth]{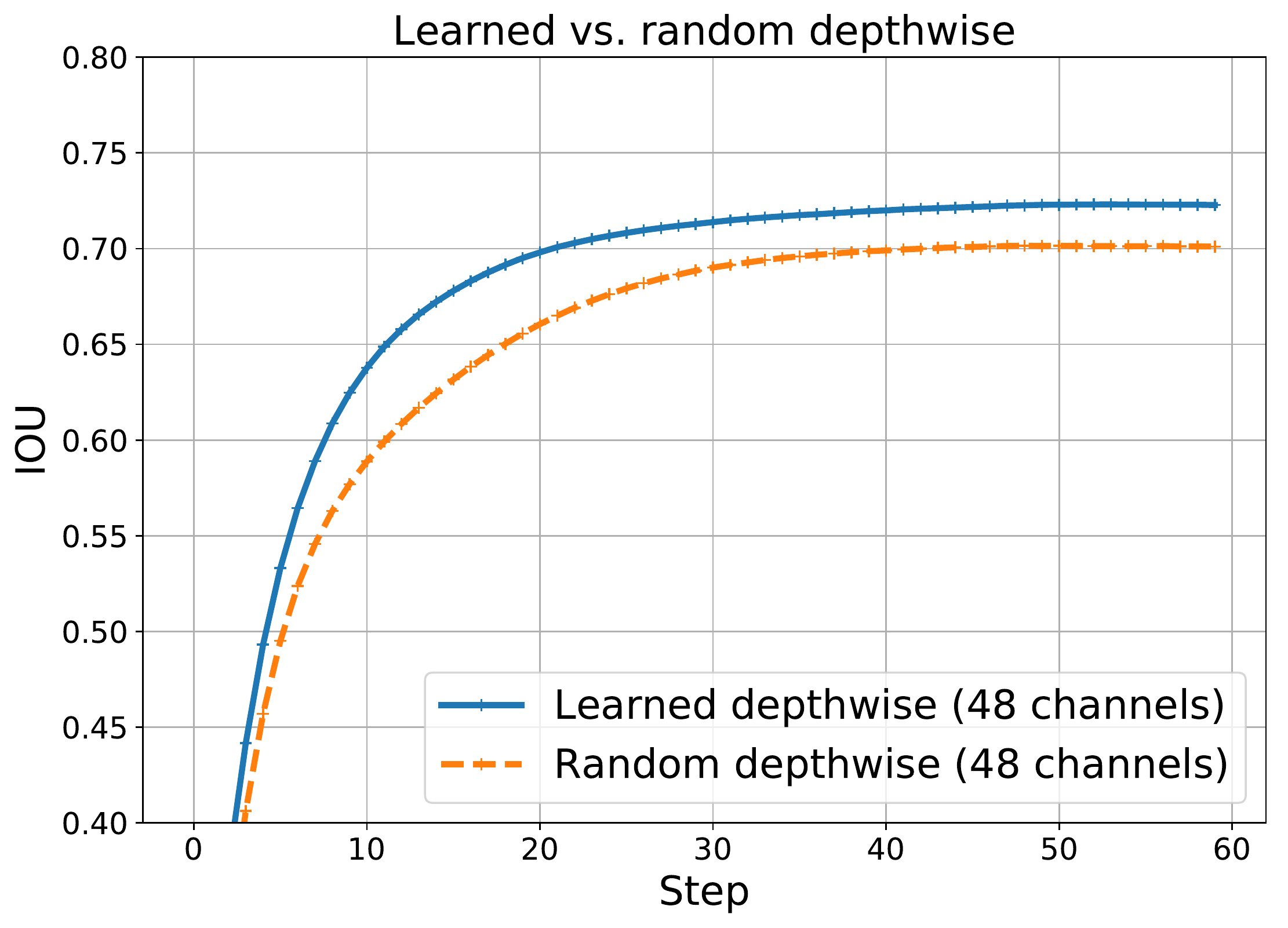}
    \caption{\small{Random vs. learned}}
    \label{fig:spatial-importance}
\end{minipage}     
\end{figure}
\begin{figure}[b]
\centering
 \begin{subfigure}[b]{0.23\textwidth}
         \centering
    \includegraphics[width=\textwidth]{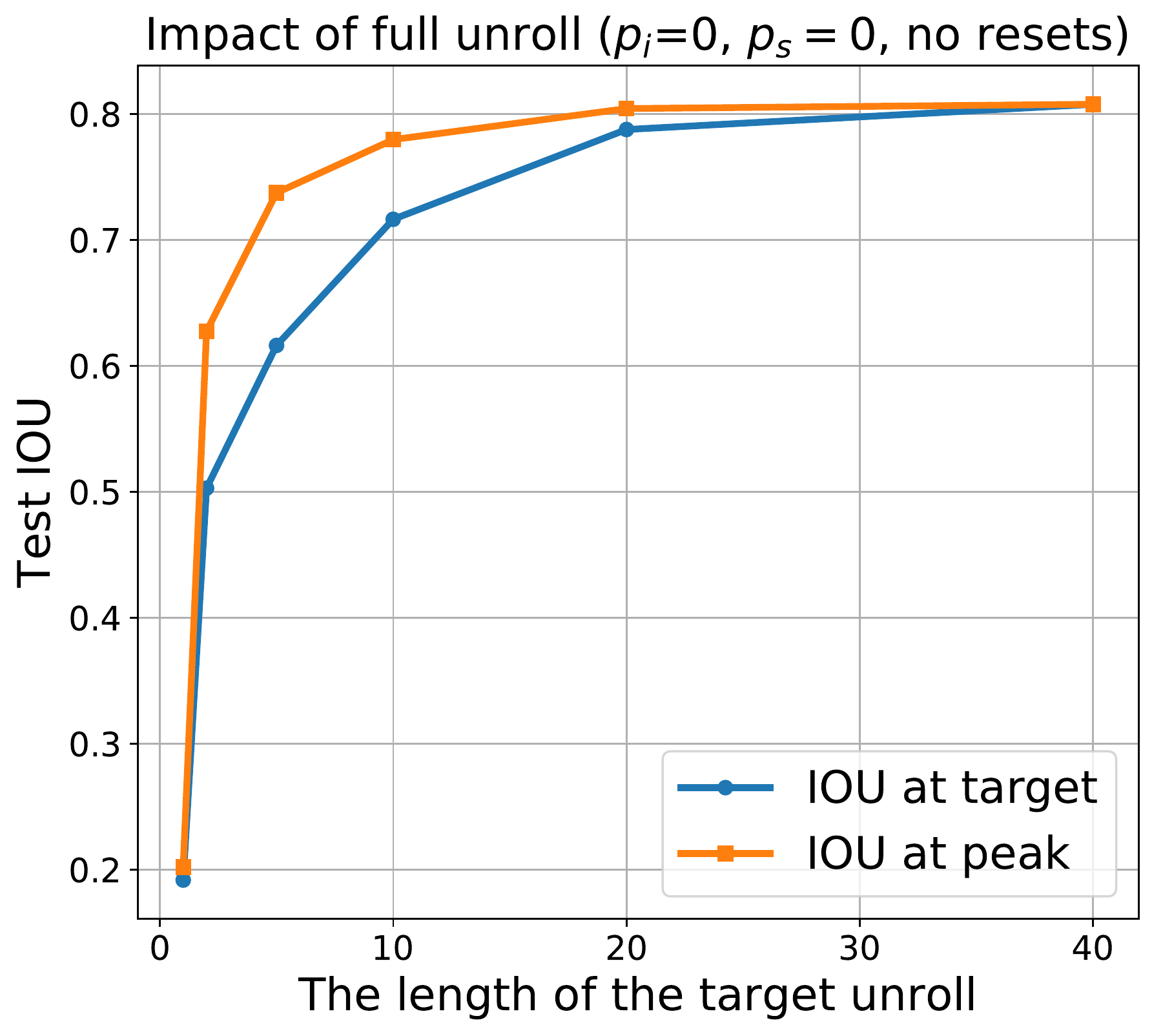}
    \caption{Classic (Full)}         \label{fig:target-is-mini-legacy}
\end{subfigure} 
 \begin{subfigure}[b]{0.23\textwidth}
         \centering
    \includegraphics[width=\textwidth]{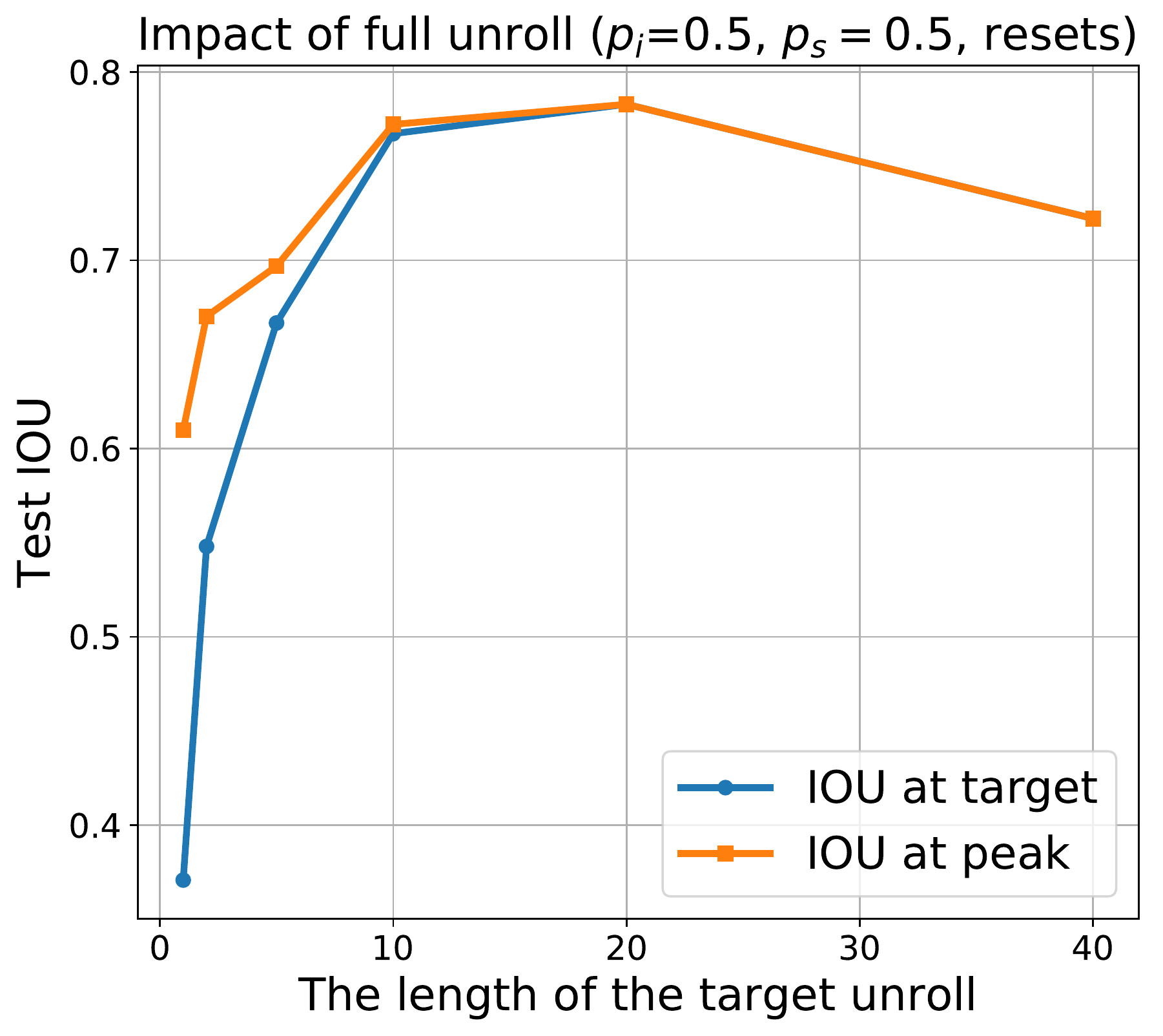}
    \caption{Mini=Target}         \label{fig:target-is-mini-stoch}
\end{subfigure} 
 \begin{subfigure}[b]{0.24\textwidth}
         \centering
    \includegraphics[width=\textwidth]{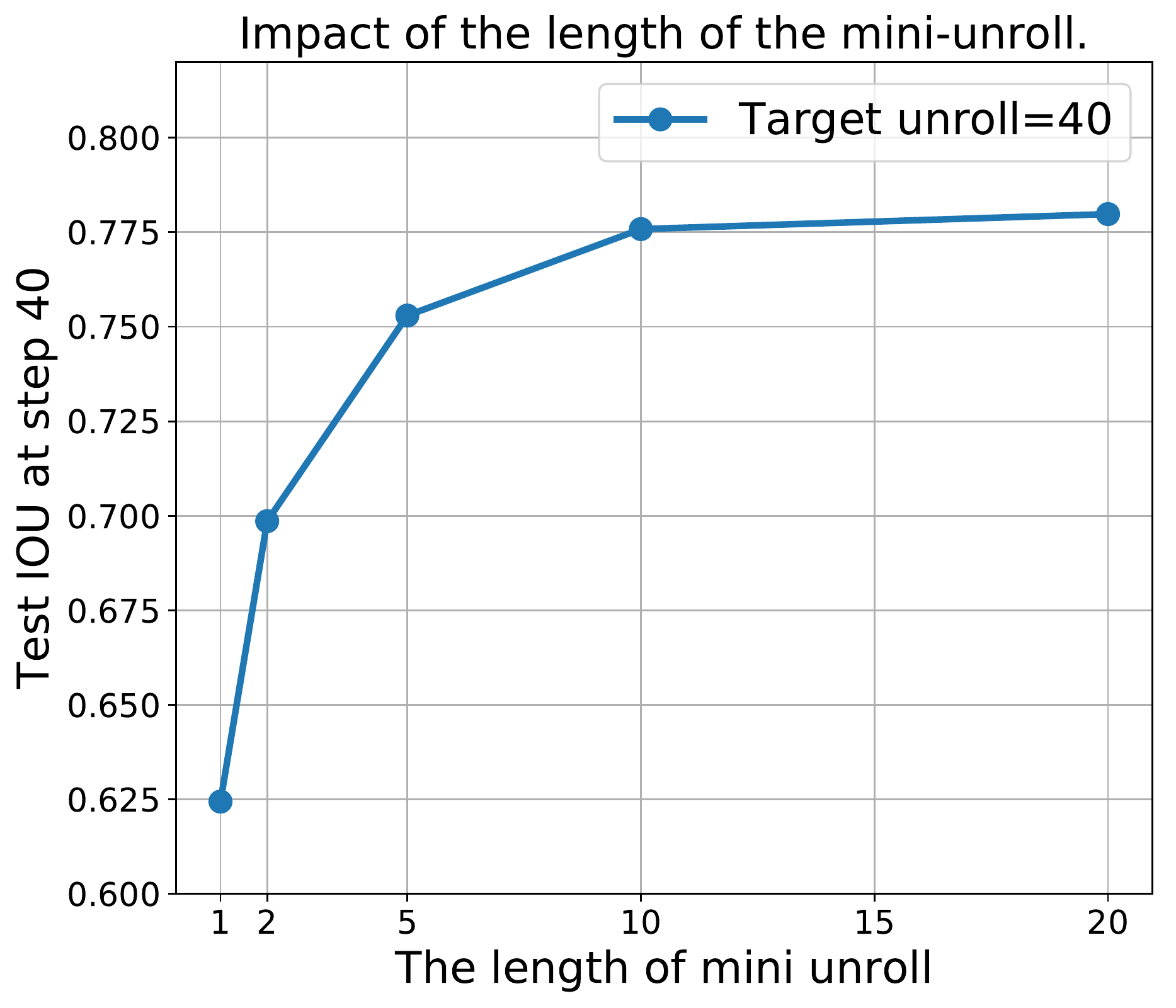}
    \caption{Target unroll 40}         \label{fig:target-unroll-40}
\end{subfigure} 
\begin{subfigure}[b]{0.24\textwidth}
\centering
    \includegraphics[width=\textwidth]{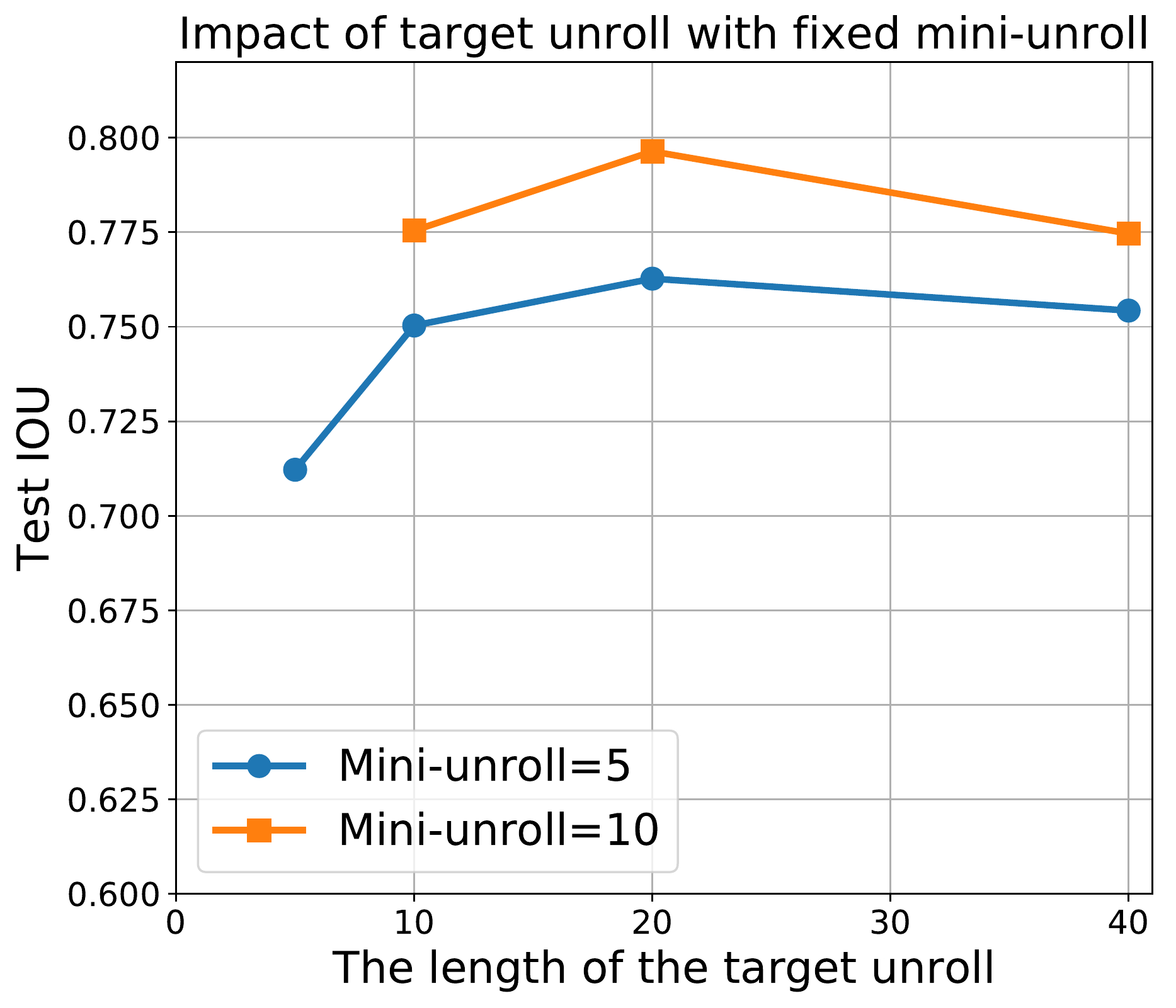}    
    \caption{Mini unroll 5/10}    \label{fig:fixed-mini-unroll}
\end{subfigure}     
\caption{\small{
    Impact of target and mini-unrolls.
    Figure~\ref{fig:target-is-mini-legacy} corresponds to a standard SGD training with no mini-unroll and no cross-step information propagation. 
    For \ref{fig:target-unroll-40} and \ref{fig:fixed-mini-unroll}, the accuracy is always measured at the $40$-step boundary.}
}
\end{figure}
\begin{figure}  
\centering
    \includegraphics[width=0.9\textwidth]{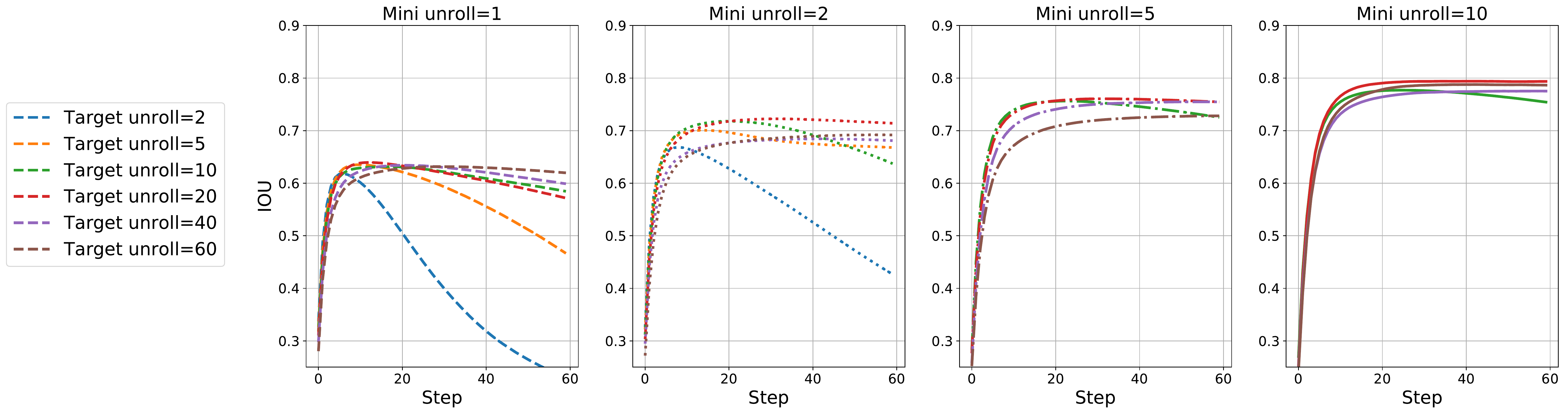}
\caption{Evolution of IOU accuracy per colony step}
\label{fig:evolution-of-iou-accuracy-for-miniunrolls}
\end{figure}

\subsection{Ablation study}
\label{sec:ablation}
\paragraph{Impact of the hidden state and the colony resolution}
    An obvious question is how important is the size of the cell state.
    Note that in our experiments we have the state size and the size of the hidden layer linked together.
    So increasing the state size also automatically increases the size of hidden state.
    It can be seen in Figure~\ref{figure:cell_state_vs_iou} that the accuracy decreases fairly quickly when the size of the state drops below $32$, and generally increases with the increased size.
    In Figure~\ref{fig:image_size_vs_iou}, we show the impact of the image resolution on the IOU.
    Here, the increased image resolution makes it harder for the colony to produce correct results.
    We show some samples of actual classifications in Figure~\ref{fig:different-resolution}.
    To address this issue for higher resolution we show some preliminary results in Section~\ref{sec:high-res}.

\paragraph{Unroll length}
    A natural question is how important is the number of steps over which the colony evolves.
    We have two parameters controlling this:
    \begin{itemize}
        \item {\em Mini unrolls} is the length of unroll that is done
    in a single step and backpropagated through and
        \item {\em  Target unroll} is the target unroll length that is used to measure loss.
    \end{itemize}
    One important consideration is that when we use mini-unroll that mismatches the target unroll, we always have to modify the training procedure to 
    preserve most of the images and cell state between training steps, because otherwise the training procedure will never observe cell states ``old enough''.
    These two hyper-parameters $p_i$ and $p_s$, are defined in section~\ref{sec:mini-unrolls}.
    For experiments with the unroll length we used $p_i=p_s=0.5$, however we note that results are not very sensitive to these parameters as long as both $p_i$ and $p_s$ are bounded away from $0$ and $1$.
    The impact of different unroll lengths and target unrolls is shown in Figure~\ref{fig:evolution-of-iou-over-long-run}.
    As can be seen from it, very short mini-unroll lengths ($<3$) lead to both very low accuracy
    and quality degradation over time, while the longer mini-unrolls and higher target unrolls lead to stable outcomes.
    All the models in this study were resettable models.
    Figure \ref{fig:evolution-of-iou-accuracy-for-miniunrolls}  shows that increasing mini-unroll essentially saturates after 10-15 steps.
    
    On the other hand, \ref{fig:target-is-mini-stoch} shows the performance over short term is best when we use monolithic training.
    While that approach results in a slightly higher accuracy, the resulting colony is susceptible to slow degradation over longer periods as shown in Figure~\ref{fig:accuracy-decay-per-step}.
    Also the importance of resettable state can be clearly seen in Figure~\ref{fig:colony_progression}, where the resettable colony does not show any degradation over long term.

\paragraph{Different normalization}
\begin{figure}[t]
\centering
 \begin{subfigure}[b]{0.4\textwidth}
         \centering
    \includegraphics[width=\textwidth]{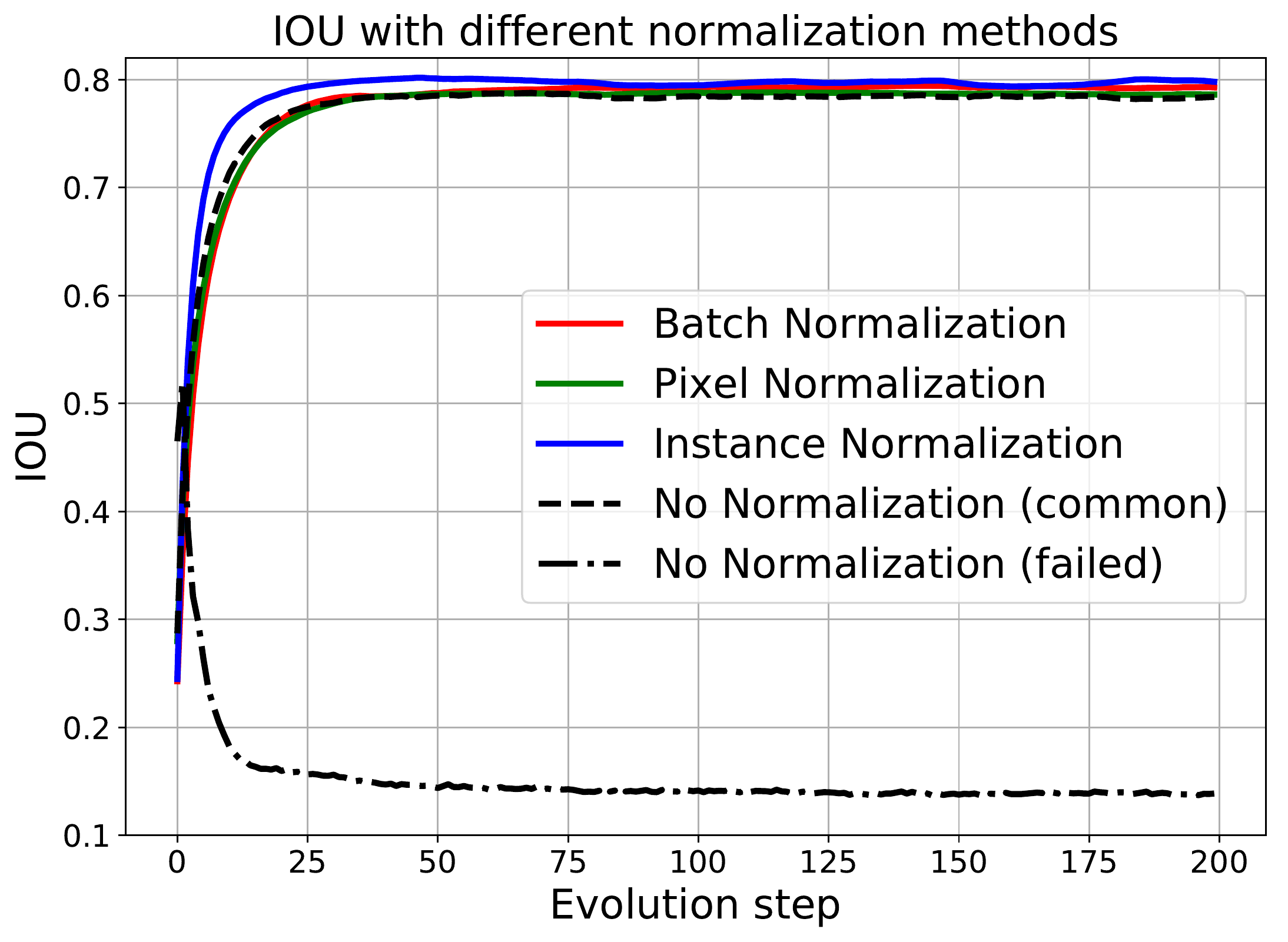}
    \caption{IOU (no error regions).}
    \label{fig:iou_different_norm_failed}
\end{subfigure} 
\begin{subfigure}[b]{0.4\textwidth}
         \centering
    \includegraphics[width=\textwidth]{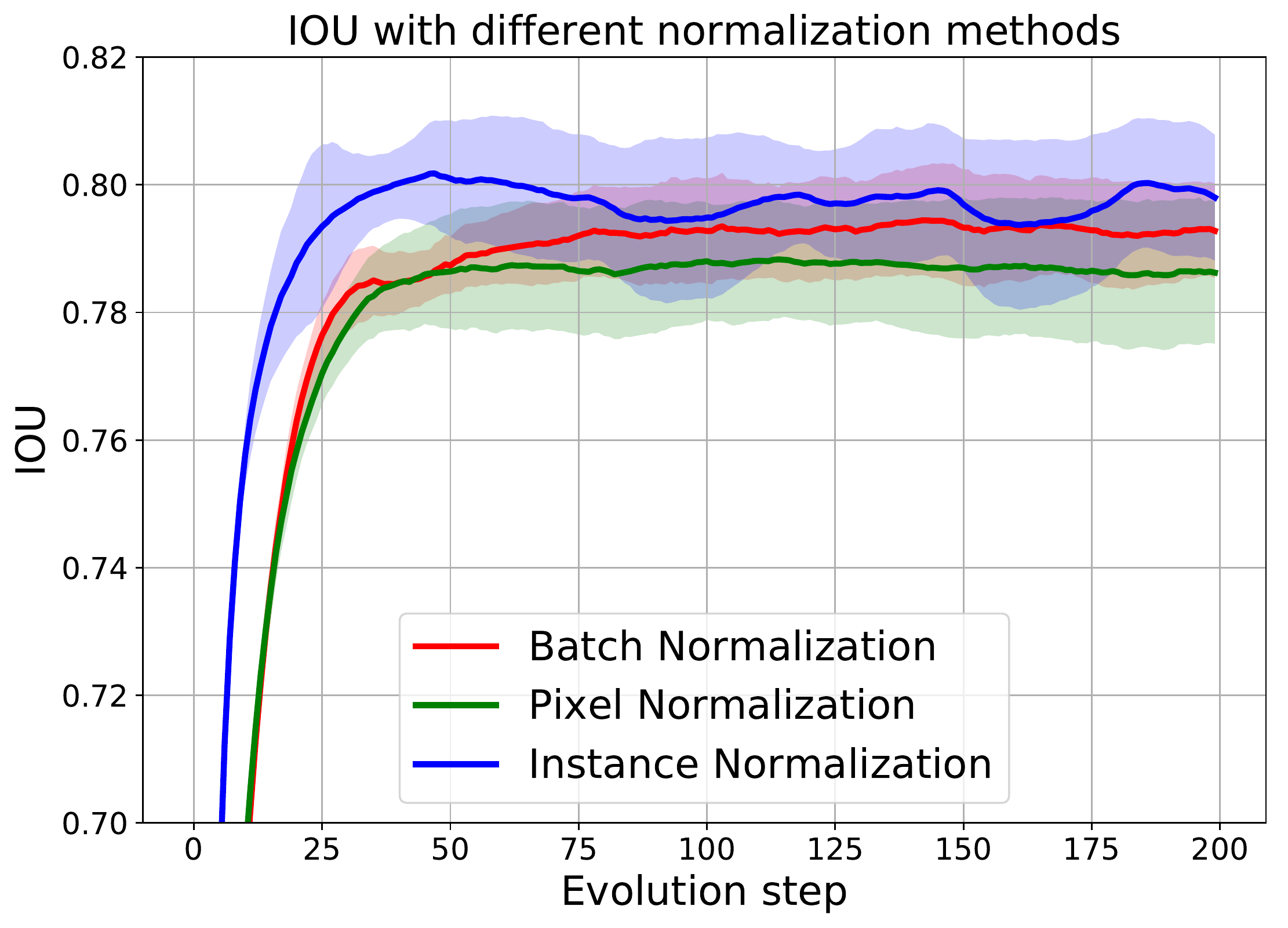}
    \caption{IOU (with error regions). }
    \label{fig:iou_different_norm_stderr}
\end{subfigure} 
\caption{\small{
    The impact of normalization methods in cellular automata. 
    Both figures plot the average IOU curve of different normalization methods on three independent runs.
    Fig.~\ref{fig:iou_different_norm_failed} includes the results on two independent runs without normalization, where one of them failed.
    Fig.~\ref{fig:iou_different_norm_stderr} shows the error regions.}}
\label{fig:iou_different_norm}
\end{figure}
    As mentioned in section~\ref{sec:state_normalization}, our models can converge well without using any normalization, while introducing it can help stabilize optimization and improve generalization ability.
    In this section we conduct experiments on using different normalization methods and illustrate the results in Fig.~\ref{fig:iou_different_norm}.
    
    We train the CA models with batch, pixel, instance and no normalization respectively, and evaluate them on the test data for multiple times.
    We can see in Fig.~\ref{fig:iou_different_norm_failed} that commonly the model trained without normalization already has good performance in terms of IOU.
    However, it may sometime fails disastrously, which indicates its instability and poor generalization.
    
    This problem can be solved by state normalization.
    Fig.~\ref{fig:iou_different_norm_stderr} shows the IOU curve using different normalization methods with error regions.
    The shaded region of each model indicates the standard deviation across its three independent runs.
    These models with normalization are much more stable than the one without, and
    their performance are relatively similar to each other regarding the major overlap among the error regions.

\paragraph{Importance of residual connections}
\begin{figure}[t]
\begin{minipage}[t]{.47\textwidth}
    \centering
        \centering
        \includegraphics[width=0.9\textwidth]{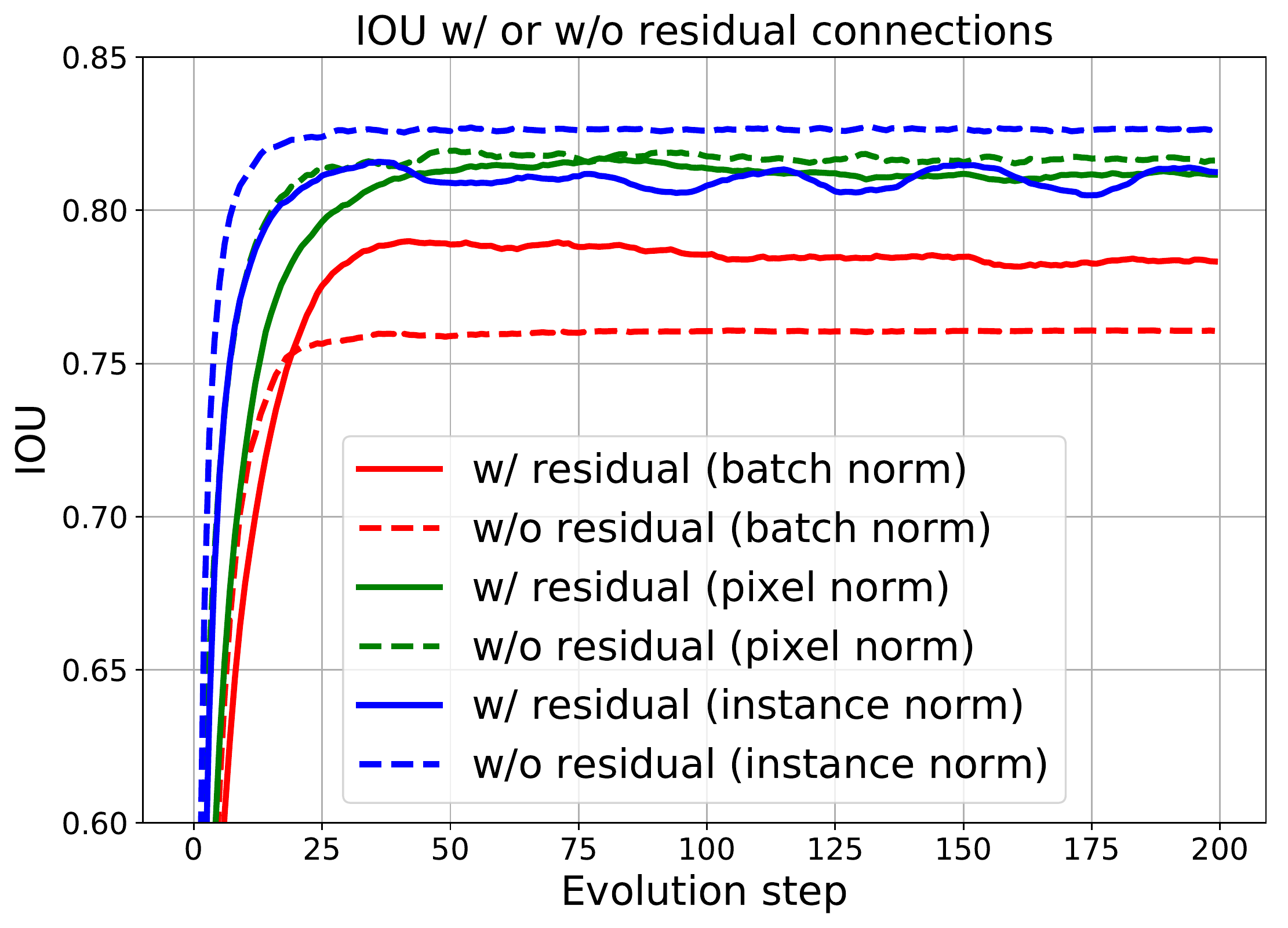}
    \caption{
        \small{The impact of residual connections. }
    }
    \label{fig:impact_of_residual_connections}
\end{minipage}
\begin{minipage}[t]{.05\textwidth}
\hspace{0.1cm}
\end{minipage}
    \begin{minipage}[t]{.47\textwidth}
    \centering
        \includegraphics[width=0.9\textwidth]{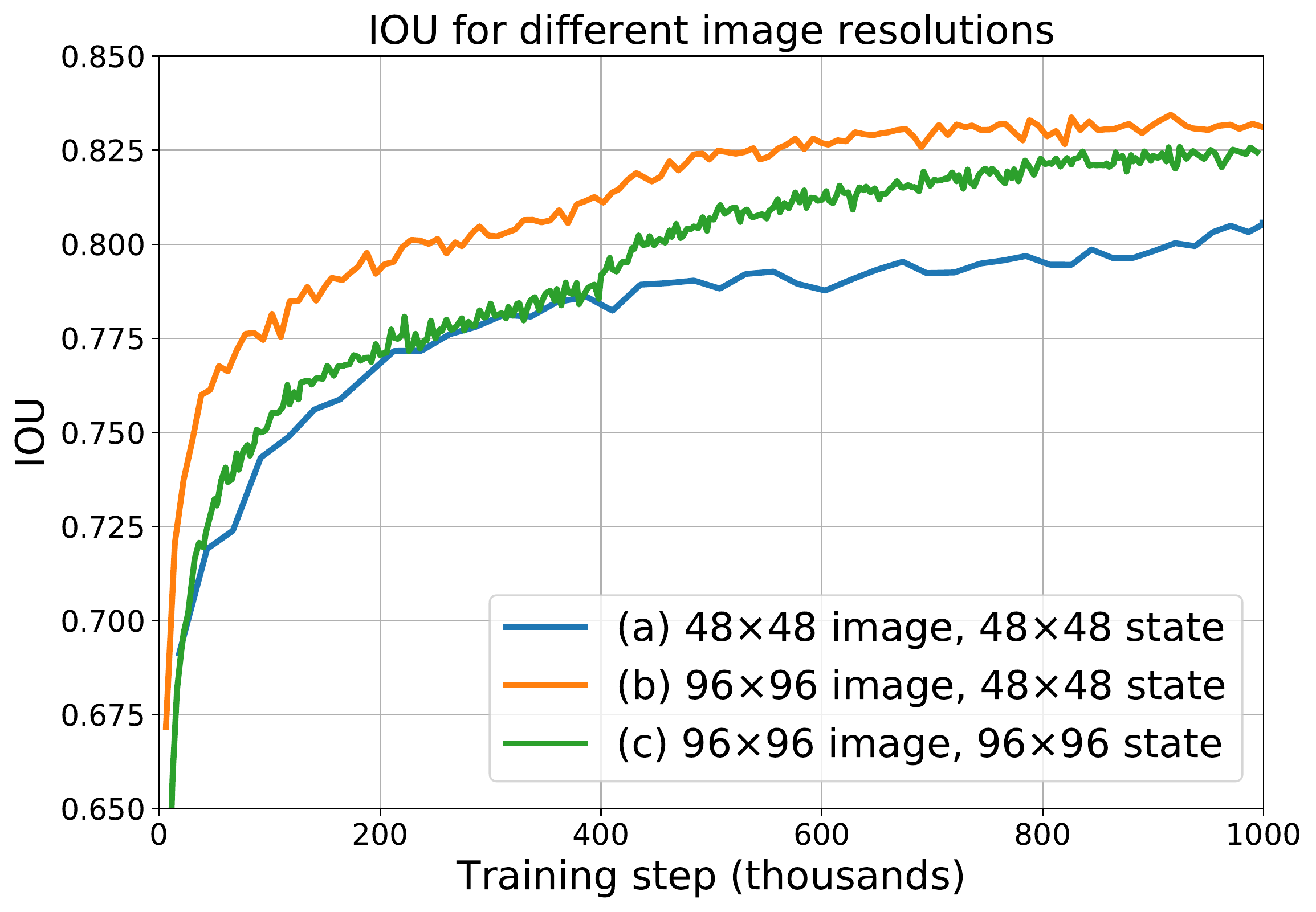}
        \caption{
                \small{(In progress.)
                 $96\times 96$ model with a $48\times 48$ internal state (see  Section~\ref{sec:high-res}) vs. original $48\times 48$ and $96\times 96$. Note: (a) is $\approx4\times $  faster than (c).}
        }
        \label{fig:iou_high_resolution}
    \end{minipage}
\end{figure}
Recall that we use a residual connection in the update rule of our cellular automation, i.e.\ $\SF(s) = U(s) + s$, for its widespread success among many applications in segmentation tasks.
But since our model is quite different from the conventional ones that we use the same rule for all the updating phases, the residual connection might behave differently from that in common models as well.
Therefore, in this section we conduct experiments to show the impact of residual connections.

To construct the model without residual connection, we use an update rule of $\SF(s) = U(s)$ where $U(s)$ has the same definition with $U(s)$ in the residual-style rule.
We run the CA model and the no-residual variant on a batch of test data and show the results in Fig.~\ref{fig:impact_of_residual_connections}.
Interestingly, unlike in other conventional models, the CA model without residual connection also has relatively good performance, and introducing the residual does not consistently guarantee an improvement in terms of IOU.
Specifically, residual connection can enhance the model with batch normalization, while the no-residual variant has similar or even slightly better performance on the model using pixel or instance normalization.
Generally speaking, our model can achieve acceptable performance in both different settings.
The reason why it can behave well without a residual-style design remains an open question and we leave it as future work.

\paragraph{Importance of learned edge detectors}
    In this section we discuss an experiment that measured the importance of the learned \emph{spatial} filters.
    It has been shown \cite{selforg} that even basic spatial filters work very well.
    We take it one step further and show that even \emph{random} filters work reasonably well, suggesting that the internal transformation of the colony is the critical component that can adopt to arbitrary spatial combination rules.
    The results are shown in Figure~\ref{fig:spatial-importance}.

\subsection{High resolution experiments}
    \label{sec:high-res}
    The approach outlined in Section~\ref{sec:high_resolution_idea} allowed us to scale our CA models to the $384\times 384$ input image size and leverage information contained in a higher-resolution image.
    In one experiment on the Oxford Cats and Dogs dataset, we compared the original CA acting on a $48\times 48$ state given a $48\times 48$ input image with another CA acting on a state of the same size, but provided with a $96\times 96$ image.
    The $96\times 96$ image was transformed with a kernel-size-3 stride-2 convolution into a $48\times 48\times 8$ input tensor that was then used by the CA.
    Similarly, each CA state was mapped into a $96\times 96$ prediction map via a kernel-size-3 stride-2 transpose convolution.
    While adding only about 3500 parameters, $96\times 96$ model improves IOU by $\approx 3\%$, see Figure~\ref{fig:iou_high_resolution}.
    Interestingly, this $96\times 96$ model also outperforms the original $96\times 96$ CA model while being almost a factor of $4$ faster.
    

\subsection{Regime change in resettable models}
\begin{figure}[t]
        \centering
        \includegraphics[width=\textwidth]{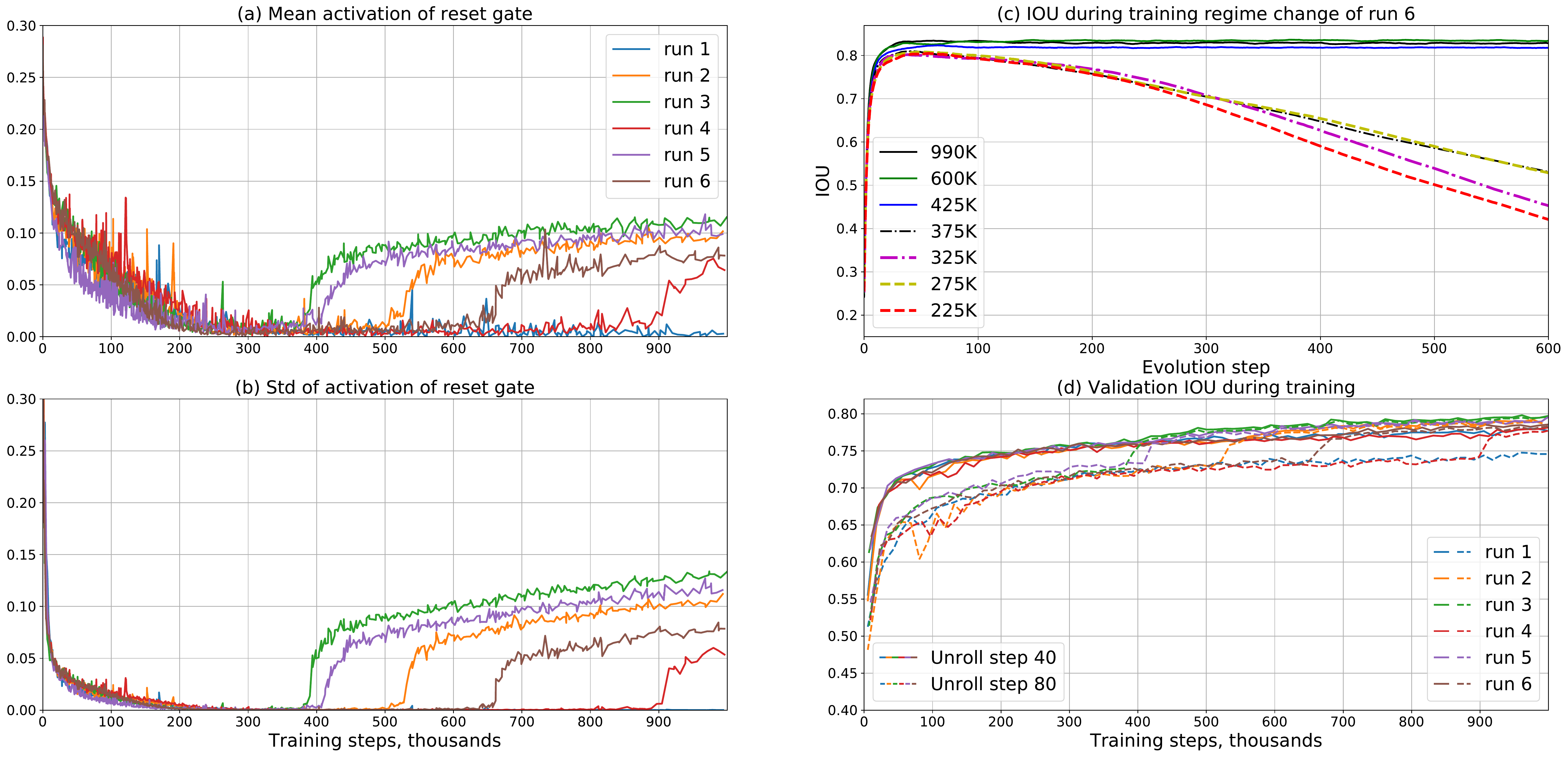}
        \caption{\small{
            Regime change of resettable gates.
            Curves in (a) and (b) correspond to gating activations in 6 identical runs (1M training steps). Figure (c) shows IOU of the run (6) at different training steps: before (dashed lines) and after the reset regime change (solid lines).
            Figure (d)  shows the accuracy transition of all runs during training. Note how the accuracy of unroll step 79, exhibits a dramatic increase once the reset gate underwent the regime change during training.}
        }
        \label{fig:regime-change}
    \end{figure}

    One interesting property that we discovered in resettable cellular automata is that they almost always undergo a regime change at some seemingly random point during training.
    In Figure~\ref{fig:regime-change}, we show how the average value of the reset gate (averaged over each pixel) changes during training.
    Notice that transition to a new regime is accompanied by a dramatic change in the model long-term stability.
    Understanding this phenomenon will be the subject of future work.


\subsection{Model size}
    One remarkable property of a cellular automaton is its size.
    For instance, our model with a hidden channel size of $48$ and $3$ layers in each update step uses only about 50K parameters.
    A modification of this cellular automaton with a first full convolutional layer replaced by a combination of a $1\times 1$ convolution and a $3x3$ depthwise convolutions has a similar performance, but uses only 18K parameters.
    Reducing the number of channels to $32$ leads to an even smaller network with $\sim$9K parameters and only moderate degradation in performance.

  In table \ref{tab:parameter-count} we present the comparative trade off accuracy numbers depending on the  size  of the model and whether we use depthwise or 3x3 convolutions.
  In this table we include both the object IOU as well as the IOU of the boundary. The depthwise version of our cellular automata has nearly 3 times fewer parameters while
  having comparable performance at many models sizes.
      \begin{table}
        \centering
        \caption{
            Model size vs. accuracy trade off. Comparison between using 1x1 + planar (depthwise) convolution vs standard 3x3 convolution as a first layer (other layers are
            always 1x1 convolutions and using different using different normalization methods. We use $(*)$ to indicate unstable runs.
        }
        \vspace{3mm}
        \begin{tabular}{|c|c|c||c|c|c||c|c|c|}
        \hline
        \multirow{2}{*}{Norm} &  \multirow{2}{*}{Cell size} & \multirow{2}{*}{Hidden size}  & \multicolumn{3}{c|}{Depthwise} &  \multicolumn{3}{c|}{$3\times3$ convolution} \\ \cline{4-9}
             &            &              &  Params & Boundary & Object   & Params & Boundary & Object \\                                      
        \hline
        %
        None   &    &     &              & 42.7     & 76  &        &  44.5 & 75 \\
        Batch  & 32 & 48  &   8.4K       & 42       & 75  &  21.4K &  43.7 & 73\\
        Inst   &    &     &              & 39       & 71  &        &  39 & 74  \\
        \hline
        None   &    &     &             & 46        &  79 &        &  46   & 77.2  \\
        Batch  & 48 & 72  &  18.3K      & 42.5      &  78 &  47K   &  46.4 & 76  \\
        Inst   &    &     &             & *    &  *       &        &  45   & 78  \\
        \hline    
        None   &    &     &             & 48        &  80 &        & 47.4  & 77.5 \\
        Batch  & 64 & 96  &  32.3K      & 46        & 79  &  82K   & 48.5  &  78 \\
        Inst   &    &     &             & *         &  *  &        & 48.6   & 80.7 \\
        \hline     
        \end{tabular}
        \label{tab:parameter-count}
    \end{table}

\section{Conclusions and Future Work}

    In this work we demonstrated that cellular automata can be trained to solve complex image segmentation tasks.
    While these automata can also be seen as neural networks, their distinctive property is the fact that each colony evolves by repeatedly applying {\em the same update rule}. 
    In order to successfully train such models, we have introduced several new techniques that stabilized the training and enabled us to use short unrolls.
    From a practical perspective, these models have multiple advantages over classical neural networks. First, they provide a very natural framework for incremental computation where underlying image can change without having to discard intermediate computations. Thus, it naturally enables us to apply segmentation on continuously changing data (such as video).
    Second, the asynchronous spatial updates they can be implemented by a grid of processors without global synchronization or global data routing.
    Third, the resulting models are remarkably compact -- less than 10K parameters for some variations. 
    
    From the theoretical perspective, we believe that cellular automata provide a promising framework, in which we can reason about how we can solve complex tasks by using multiple simple independent agents. For example our CA design can easily generalize to allow for richer interactions. For instance cells can be allowed to copy themselves to move around the grid, or leave breadcrumbs for other cells to utilize. Training such CA remains subject of further work.

\clearpage

%
%
\bibliographystyle{splncs04}
\bibliography{main}
\appendix
\section{Appendix}

\subsection{Empirical differences between resettable and non-resettable models}
    
\begin{figure}
    \centering
    \begin{subfigure}[c]{0.15\textwidth}
    \vspace{1.25cm}
        \includegraphics[width=.8\textwidth]{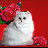}
        \vspace{1.25cm}
        \caption{Input}
        \label{fig:ev_image}
    \end{subfigure}    
    \begin{subfigure}[c]{0.4\textwidth}
        \centering
        \includegraphics[width=.9\textwidth]{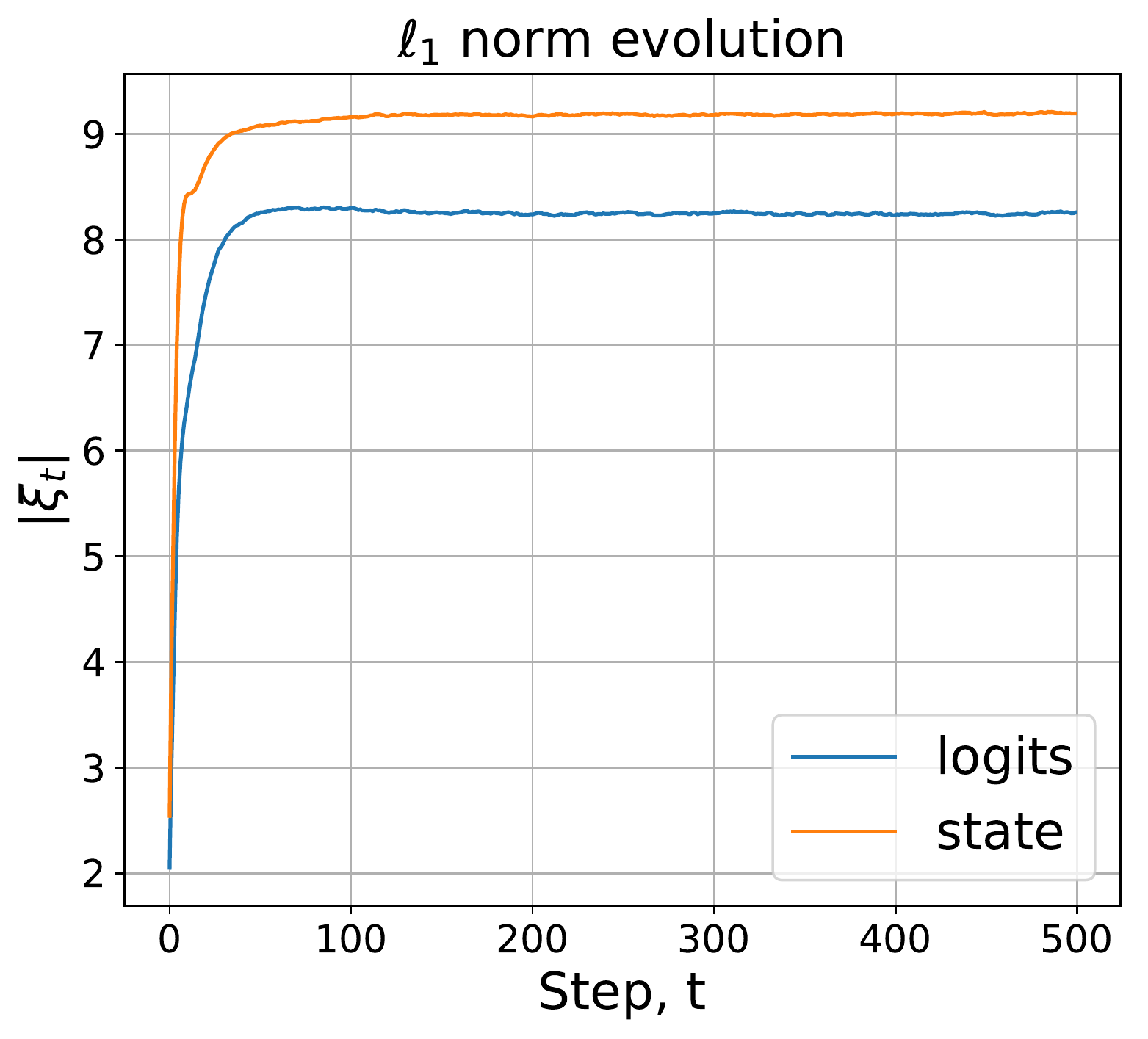}
        \caption{Resettable model}
    \end{subfigure}
    \begin{subfigure}[c]{0.4\textwidth}
        \centering
        \includegraphics[width=.9\textwidth]{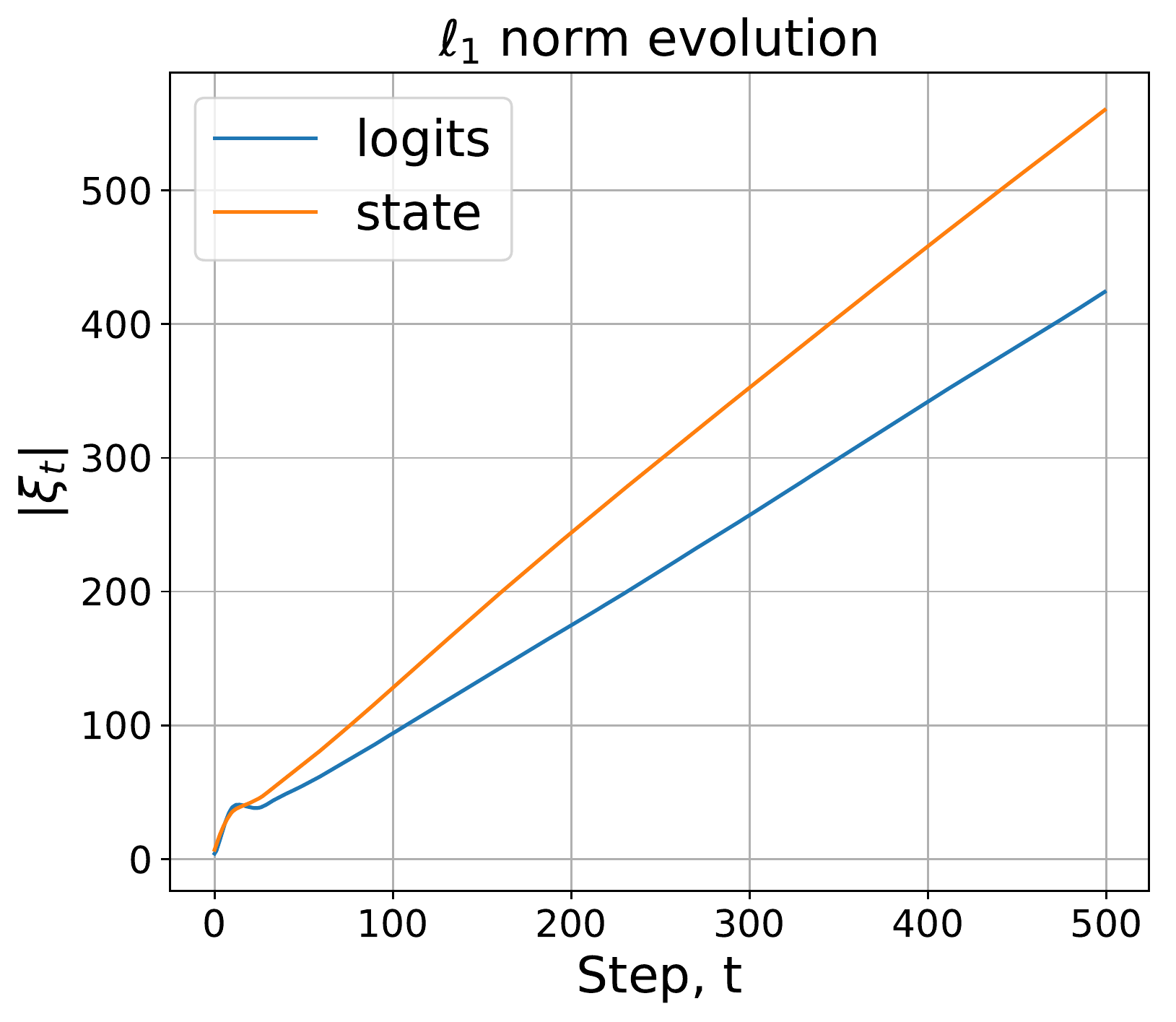}
        \caption{Non-resettable model}
    \end{subfigure}
    \caption{\small{
        Evolution of the $\ell_1$ norm of the CA state and logits for the input image (a) for models with instance normalization.}
    }
    \label{fig:norm}
    \end{figure}

\begin{figure}
    \centering
    \begin{subfigure}[c]{0.45\textwidth}
        \centering
        \includegraphics[width=.9\textwidth]{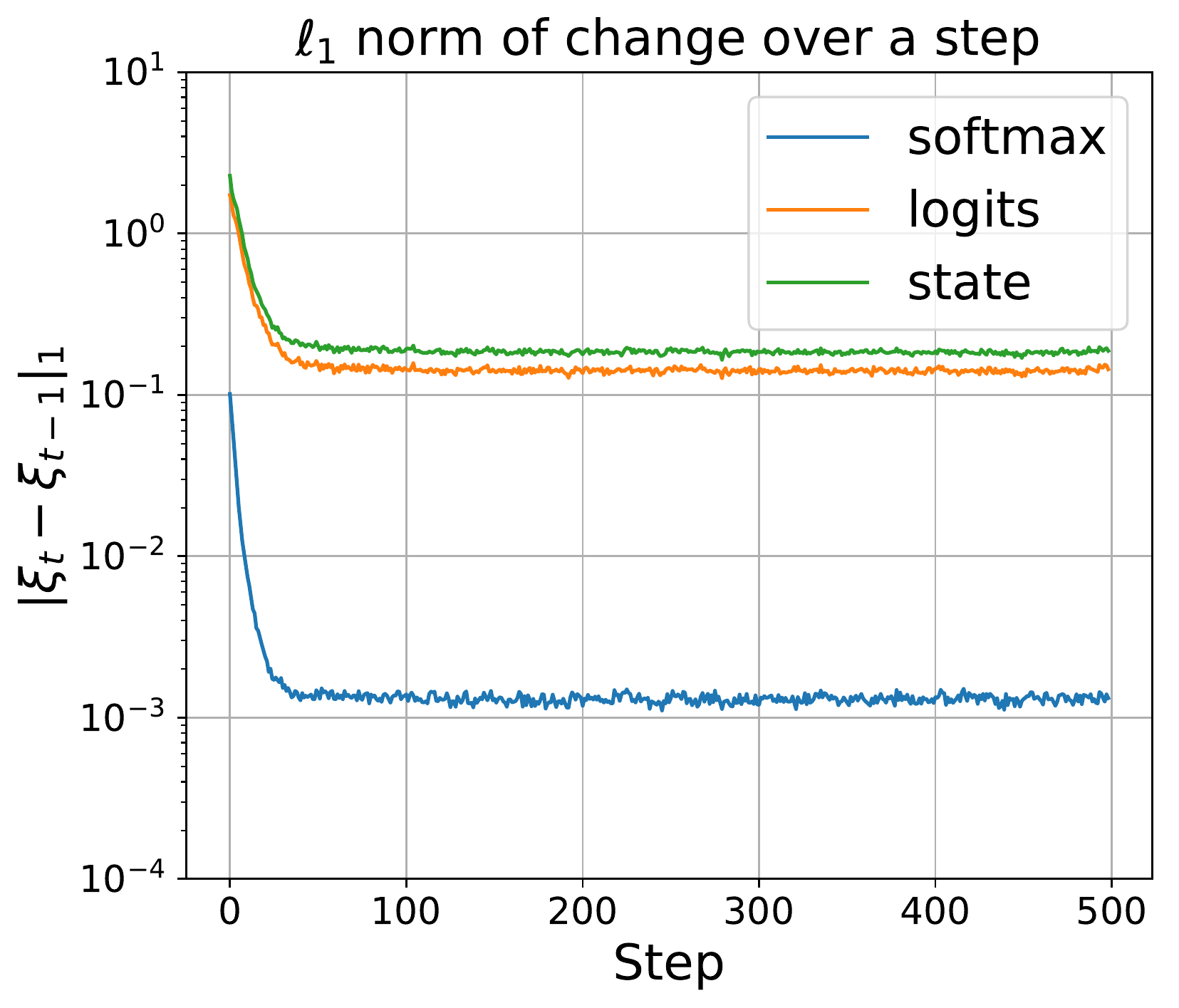}
        \caption{Resettable model}
    \end{subfigure}
    \begin{subfigure}[c]{0.45\textwidth}
        \centering
        \includegraphics[width=.9\textwidth]{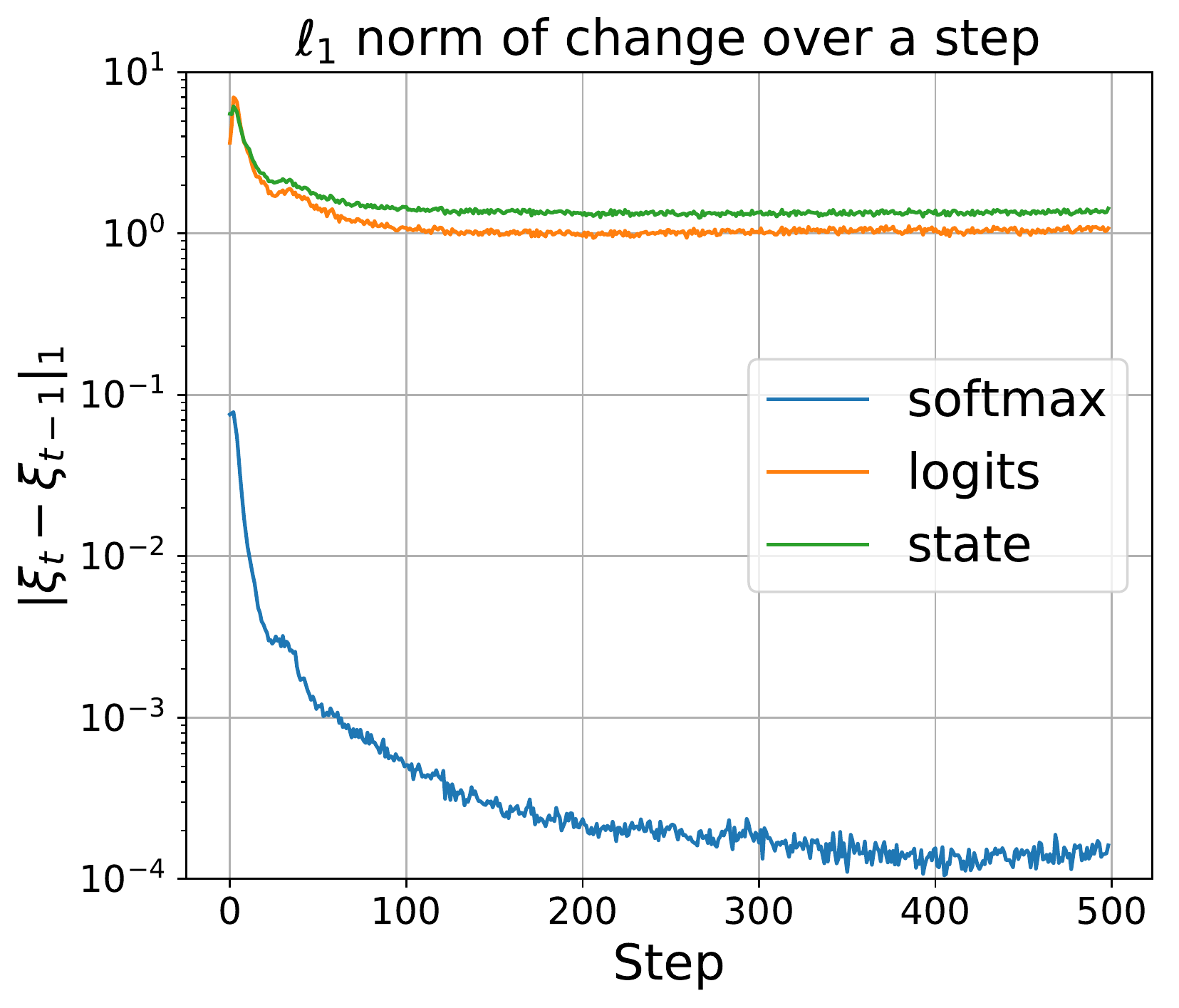}
        \caption{Non-resettable model}
    \end{subfigure}
    \caption{\small{
        Evolution of the $\ell_1$ norm of a single-step change for the hidden CA state, logits and final predictions for models with instance normalization. 
        The input image is shown in Figure~\ref{fig:ev_image}.}
    }
    \label{fig:delta}
    \end{figure}

    As discussed in the main text, resettable models produce stable predictions that do not change significantly even if the cellular automaton (CA) is evolved for hundreds of steps beyond the target.
    It turns out that these two types of models exhibit other dramatically different behaviors.
    For example, as shown in Figure~\ref{fig:norm}, $\ell_1$ norms of the CA hidden state and model logits saturate in a resettable model with {\em instance normalization}, but grow linearly with the step number in a non-resettable one.
    The linear growth in a model with instance normalization is possible due to the presence of residual connections.
    Norms shown in Figure~\ref{fig:norm} are normalized to the total number of dimensions in each variable.
    Looking at the step-by-step changes, we see that fluctuations of the hidden states and logits saturate in amplitude for both models (see Figure~\ref{fig:delta}).
    However, since the norm of the logits grows linearly in time, final predictions decay in a non-resettable model (while saturating in the resettable CA).

\begin{figure}
    \centering
    \begin{subfigure}[c]{0.4\textwidth}
        \centering
        \includegraphics[width=.9\textwidth]{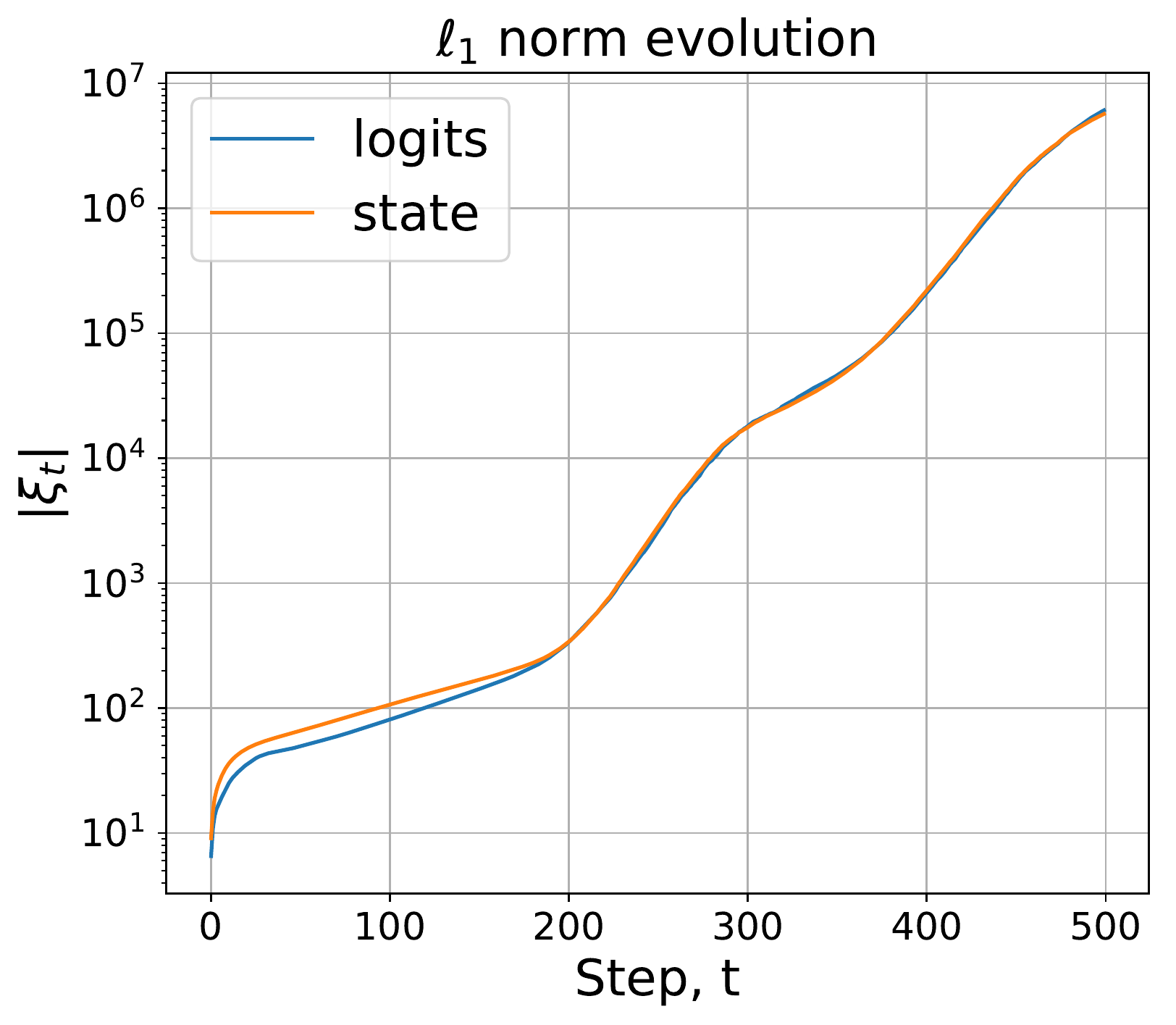}
        \caption{Non-resettable model}
    \end{subfigure}
    \begin{subfigure}[c]{0.4\textwidth}
        \centering
        \includegraphics[width=.9\textwidth]{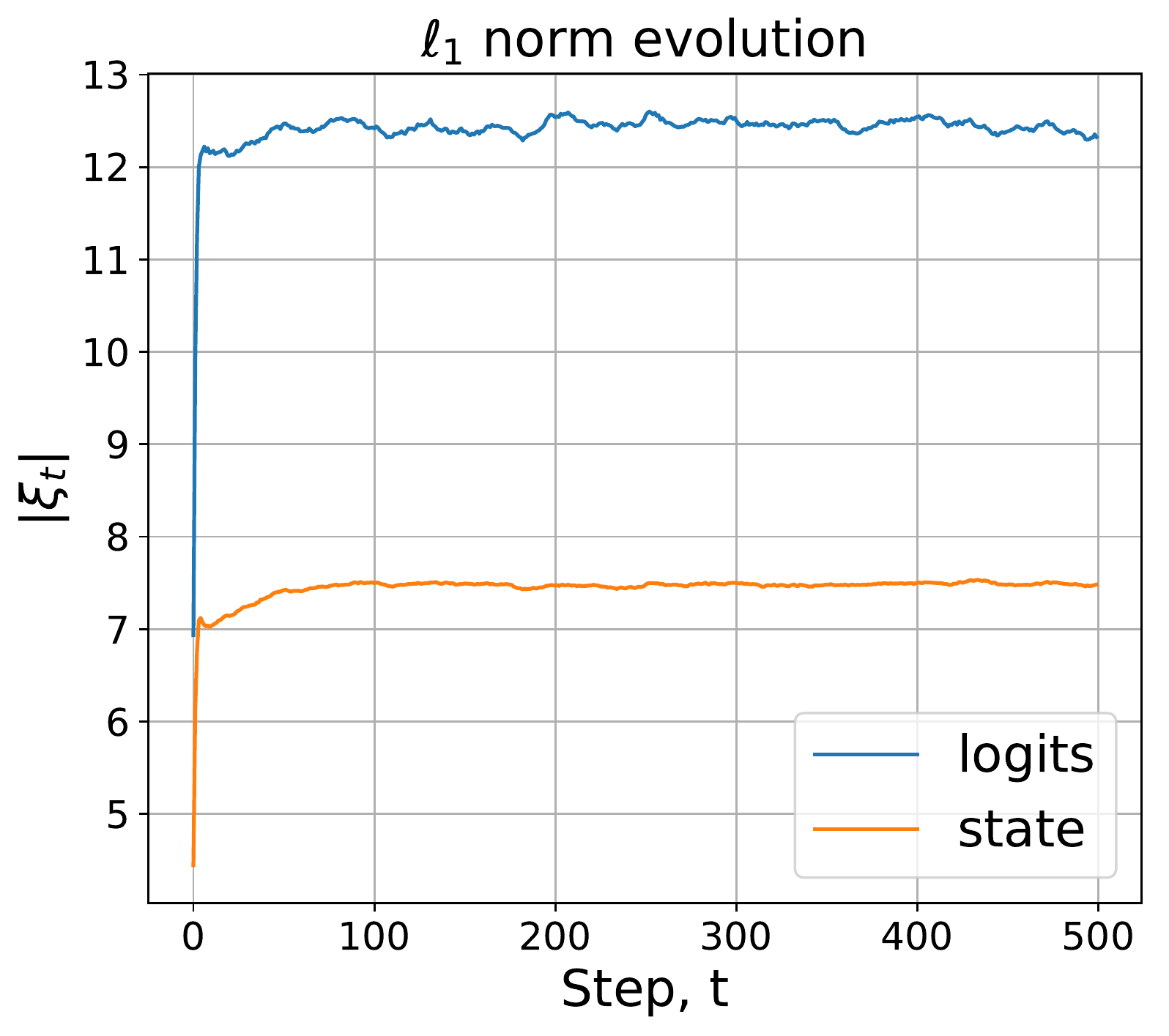}
        \caption{Resettable model}
    \end{subfigure}
    \caption{\small{
        Evolution of the $\ell_1$ norm of the CA state and logits for models without any normalization. 
        The input image is shown in Figure~\ref{fig:ev_image}.
    }}
    \label{fig:nonorm_noreset}
    \end{figure}

\begin{figure}
    \centering
    \begin{subfigure}[c]{0.4\textwidth}
        \centering
        \includegraphics[width=.9\textwidth]{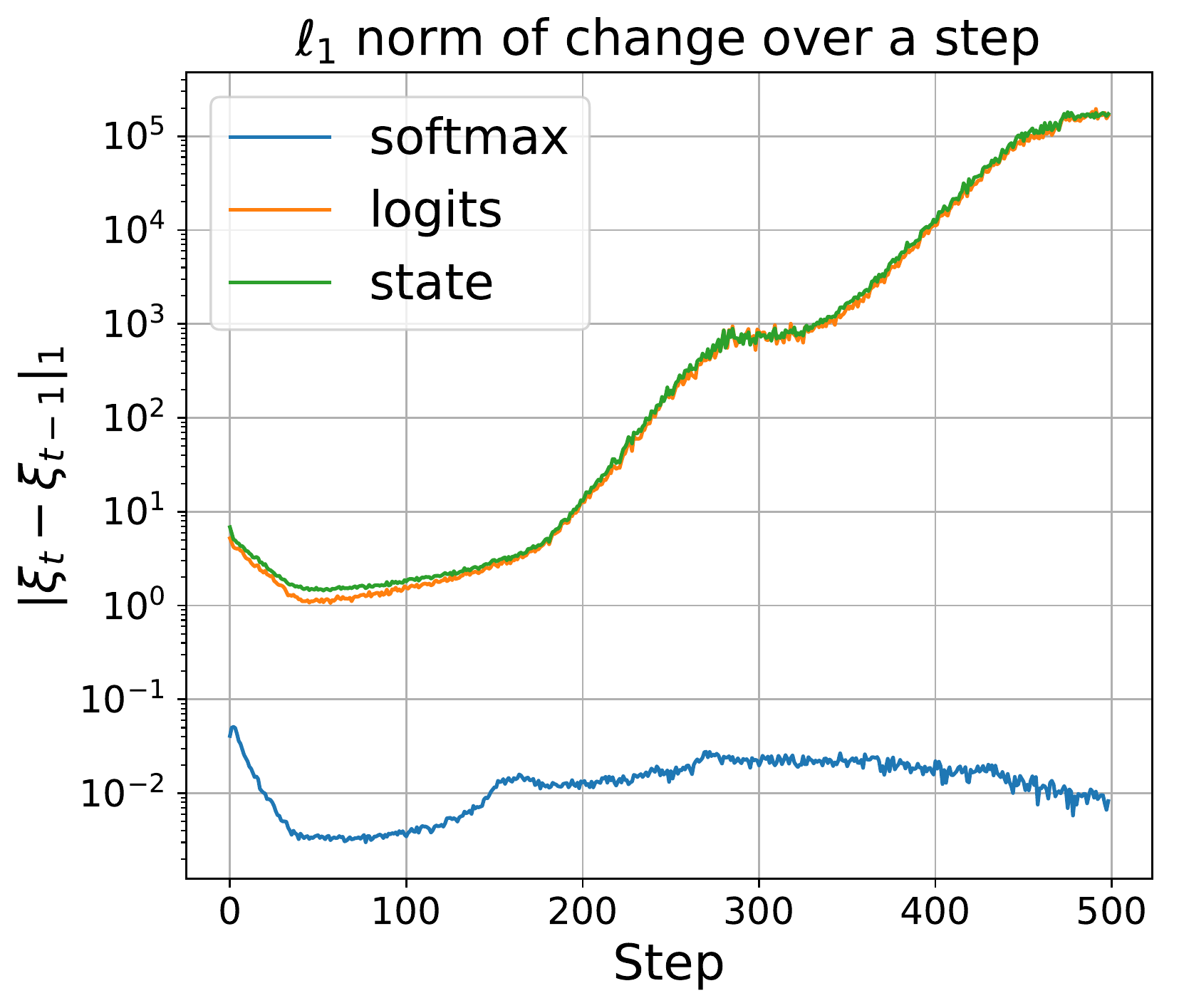}
        \caption{Non-resettable model}
    \end{subfigure}    
    \begin{subfigure}[c]{0.4\textwidth}
        \centering
        \includegraphics[width=.9\textwidth]{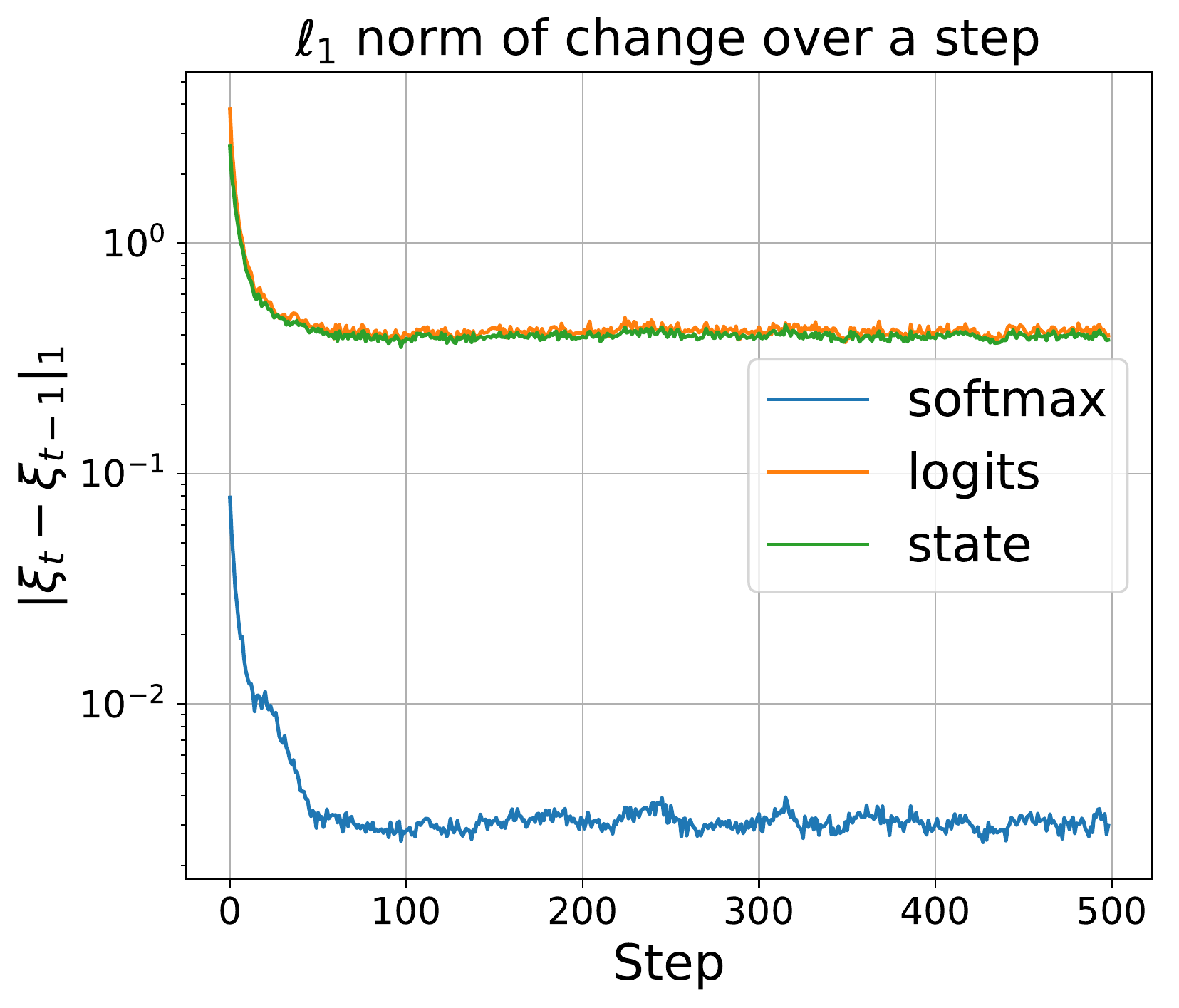}
        \caption{Resettable model}
    \end{subfigure}
    \caption{\small{
        Evolution of the $\ell_1$ norm of a single-step change of the CA state, logits and predictions for models without normalization. 
        The input image is shown in Figure~\ref{fig:ev_image}.
    }}
    \label{fig:nonorm_reset}
    \end{figure}

    The difference between resettable and non-resettable models is even more striking in models that do not use normalization.
    As can be seen in Figures~\ref{fig:nonorm_noreset} and \ref{fig:nonorm_reset}, non-resettable models exhibit exponential hidden state growth when unrolled over periods $10\times$ their target unroll.
    On the other hand, resettable models learn to balance CA hidden state and logit $\ell_1$ norms at remarkably stable levels both with and \emph{without} normalization, even though they were not specifically trained for it.

 \subsection{Adversarial images}
    
\begin{figure}
    \centering
    \begin{subfigure}[c]{0.24\textwidth}
        \centering
        \includegraphics[width=.94\textwidth]{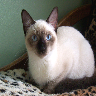}
        \includegraphics[width=.94\textwidth]{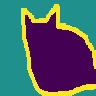}
        \caption{\small{Original}}
        \label{fig:adv_original}
    \end{subfigure}
    \begin{subfigure}[c]{0.24\textwidth}
        \centering
        \includegraphics[width=.94\textwidth]{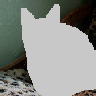}
        \includegraphics[width=.94\textwidth]{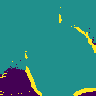}
        \caption{\small{Grayed-out}}
    \end{subfigure}
    \begin{subfigure}[c]{0.24\textwidth}
        \centering
        \includegraphics[width=.94\textwidth]{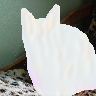}
        \includegraphics[width=.94\textwidth]{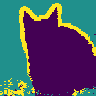}
        \caption{\small{Adversarial}}
    \end{subfigure}
    \begin{subfigure}[c]{0.24\textwidth}
        \centering
        \includegraphics[width=.9\textwidth]{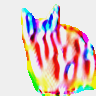}
        \caption{\small{Perturbation}}
    \end{subfigure}
    \caption{\small{
        {\em An adversarial image for a pre-trained $96\times 96$ resettable CA.}
        We start with an original image shown in (a) together with the CA prediction.
        We then cut out a pet region and replace it with a homogeneous gray color as shown in (b).
        We then use a deepdream technique to find a perturbation localized within the grayed-out pet region that tricks CA into classifying this image region as a ``pet''.
        The resulting adversarial image and the CA prediction for it are shown in (c).
        Notice that by modifying the region inside the animal outline, we can also affect pixel classification outside of it.
        Image (d) shows an amplified adversarial perturbation.
        }
    }
    \label{fig:deepdream}
    \end{figure}

    Like the vast majority of other deep learning models, our cellular automata are susceptible to adversarial attacks.
    Images tricking CA into improperly classifying certain regions can be found using a conventional deepdream technique.
    For example, in Figure~\ref{fig:deepdream}, we show an image optimized to trick a pre-trained CA into classifying a large image region as pet while it is clear that there is no pet in the image.

 \subsection{High resolution experiments}
 
\begin{figure}
    \centering
    \begin{subfigure}[c]{0.32\textwidth}
        \centering
        \includegraphics[width=.98\textwidth]{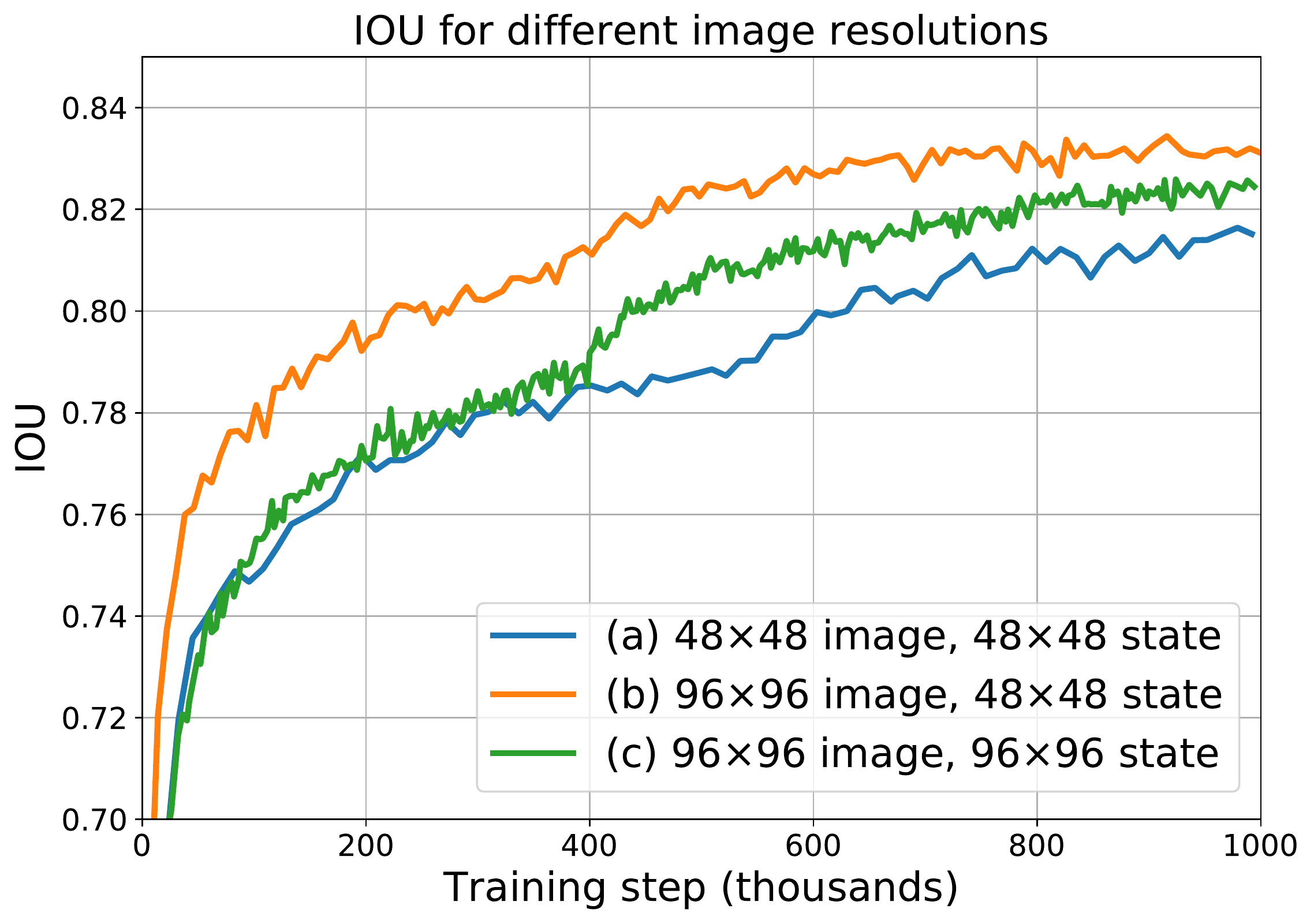}
        \caption{``object''}
    \end{subfigure}
    \begin{subfigure}[c]{0.32\textwidth}
        \centering
        \includegraphics[width=.98\textwidth]{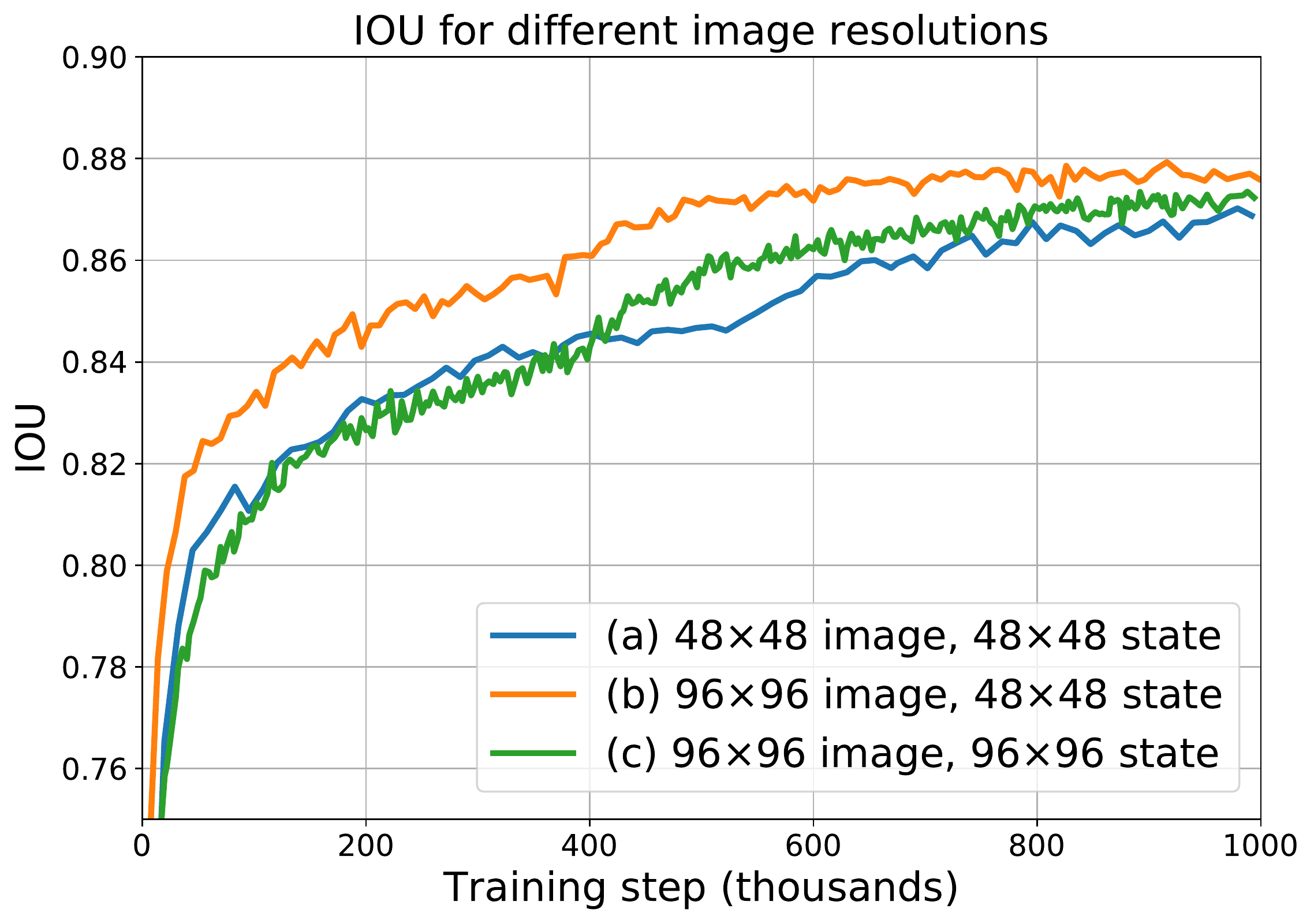}
        \caption{``background''}
    \end{subfigure}
    \begin{subfigure}[c]{0.32\textwidth}
        \centering
        \includegraphics[width=.98\textwidth]{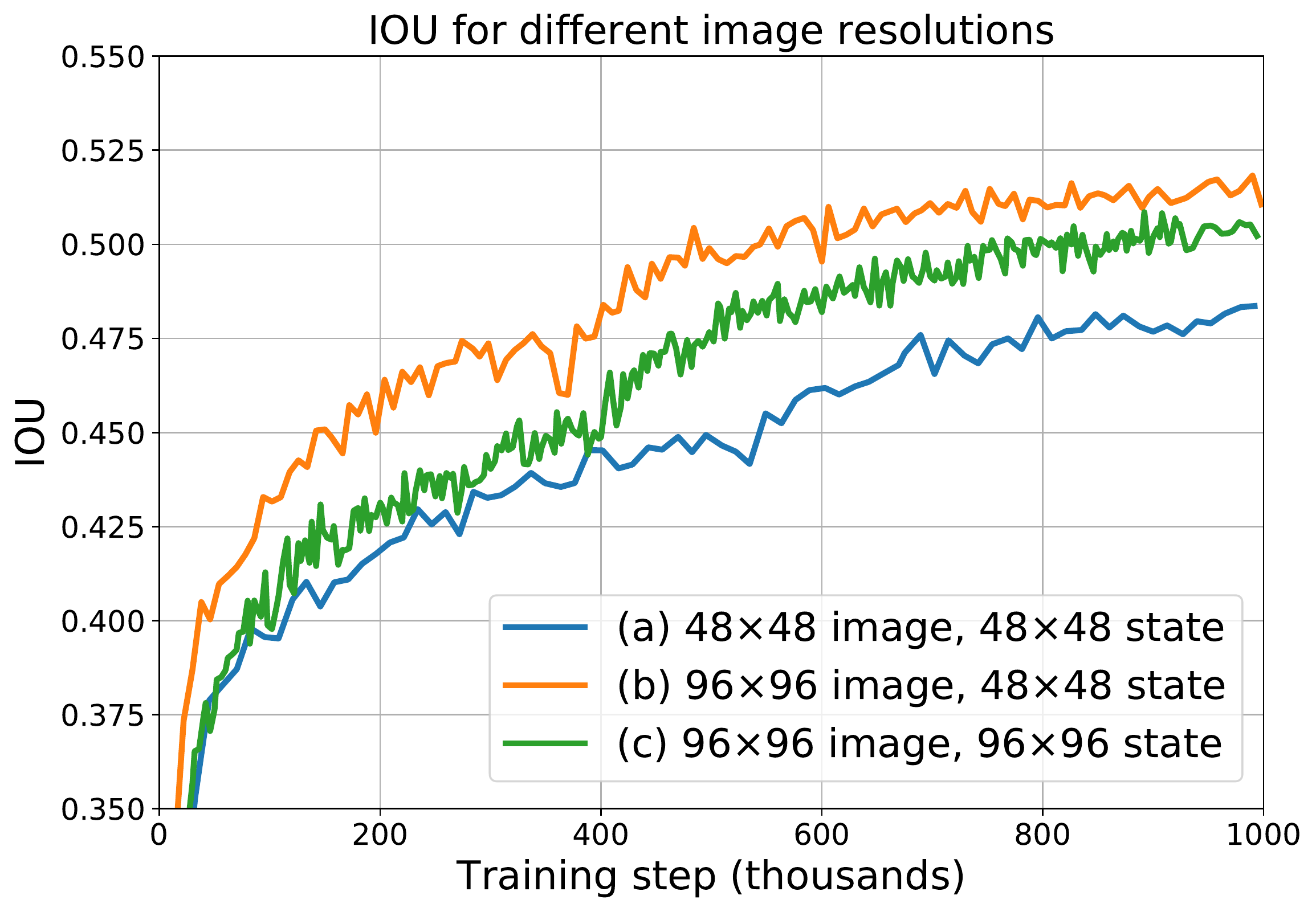}
        \caption{``boundary''}
    \end{subfigure}
    \caption{\small{
        Training curves showing IOU for three labels (``object'', ``background'' and ``boundary'').
        We compare three models: (i) $48\times 48$ image resolution with $48\times 48$ state resolution; (ii) $96\times 96$ image resolution with $96\times 96$ state resolution; (iii) $96\times 96$ image resolution with $48\times 48$ state resolution.
        Notice that $96\times 96$ model with $48\times 48$ state resolution is about 4 times faster than the model with $96\times 96$ state resolution.
        }
    }
    \label{fig:resolutions}
    \end{figure}

    In our experiments in the main paper on the Oxford Cats and Dogs dataset, we compared cellular automata running on $48\times 48$ and $96\times 96$ images using the same CA state size with the CA processing  $96\times 96$ image, but using a $48\times 48$ hidden state resolution.
    Here we show results of completed runs. 
    Figure~\ref{fig:resolutions} shows training curves for these three models.
    The model acting on a $96\times 96$ image, but using a $48\times 48$ hidden state resolution outperformed all other models with respect to all 3 labels.
    
\end{document}